%% file: main.tex
\documentclass[runningheads]{llncs}
\usepackage[T1]{fontenc}
\usepackage{graphicx}

\input{preamble}

\usepackage{color}
\usepackage{hyperref}
\usepackage[capitalize]{cleveref}

\usepackage[numbers]{natbib} %
\usepackage{booktabs}
\newcommand{\ie}{i.e.\ }

\usepackage{capt-of}
\usepackage{wrapfig}

\crefname{section}{Sec.}{Secs.}
\Crefname{section}{Sec.}{Secs.}
\crefname{table}{Tab.}{Tabs.}
\Crefname{table}{Tab.}{Tabs.}

\begin{document}
\title{
  Yesnt: Are Diffusion Relighting Models Ready for Capture Stage Compositing?\\
  A Hybrid Alternative to Bridge the Gap
}
\titlerunning{
  Are Diffusion Relighting Models Ready for Capture Stage Compositing?
}

\author{
Elisabeth J{\"u}ttner \and
Janelle Pfeifer \and
Leona Krath \and
Stefan Korfhage \and \\
Hannah Dr{\"o}ge \and
Matthias B. Hullin \and
Markus Plack
}

\authorrunning{E. J{\"u}ttner et al.}

\institute{
University of Bonn, Bonn, Germany\\
\email{\{elisabeth.juettner,leona,stefan\}@uni-bonn.de}\\
\email{\{pfeifer,droege,hullin,plack\}@cs.uni-bonn.de}
}

\maketitle              %

\input{sec/0_abstract}

\input{sec/1_intro}
\input{sec/2_rel_work}

\input{sec/3_method}
\input{sec/4_evaluation}

\input{sec/5_conclusion}

\bibliographystyle{splncs04}
\bibliography{main}

\end{document}

%% file: preamble.tex
\usepackage{amsmath}      %
\usepackage{amssymb}      %
\usepackage{bm}           %
\usepackage{graphicx}     %
\usepackage{mathtools}    %
\usepackage{bbm}          %

\usepackage[dvipsnames]{xcolor}

\newcommand{\HD}[1]{\textcolor{red}{HD: #1}}
\newcommand{\JP}[1]{\textcolor{olive}{JP: #1}}
\newcommand{\EJ}[1]{\textcolor{yellow}{EJ: #1}}
\newcommand{\newp}[1]{\textcolor{magenta}{#1}}
\newcommand{\delete}[1]{}

%% file: sec/0_abstract.tex
\begin{abstract}
Volumetric video relighting is essential for bringing captured performances into virtual worlds, but current approaches struggle to deliver temporally stable, production-ready results. Diffusion-based intrinsic decomposition methods show promise for single frames, yet suffer from stochastic noise and instability when extended to sequences, while video diffusion models remain constrained by memory and scale.

We propose a hybrid relighting framework that combines diffusion-derived material priors with temporal regularization and physically motivated rendering. Our method aggregates multiple stochastic estimates of per-frame material properties into temporally consistent shading components, using optical-flow-guided regularization. For indirect effects such as shadows and reflections, we extract a mesh proxy from Gaussian Opacity Fields and render it within a standard graphics pipeline.
Experiments on real and synthetic captures show that this hybrid strategy achieves substantially more stable relighting across sequences than diffusion-only baselines, while scaling beyond the clip lengths feasible for video diffusion. These results indicate that hybrid approaches, which balance learned priors with physically grounded constraints, are a practical step toward production-ready volumetric video relighting. 

{\small\normalfont\raggedleft
\vspace{0.2cm}Project Page: \url{https://ejuet.github.io/yesnt/}\par}
\vspace{-0.5\baselineskip}
\end{abstract}

%% file: sec/1_intro.tex
\input{figures/teaser/teaser}

\section{Introduction}
\label{sec:intro}

Volumetric capture stages provide a compelling platform for recording human performances in 3D, enabling applications such as immersive telepresence, virtual production, and mixed-reality content creation. These systems allow for captured performances to be replayed from arbitrary viewpoints and inserted into novel environments. Realizing this potential, however, requires addressing two fundamental challenges: \emph{novel view synthesis}, to render unseen perspectives from sparse camera inputs, and \emph{relighting}, to adapt the captured content to new lighting conditions so that it blends naturally into target scenes.

The first challenge, novel view synthesis (NVS), has seen rapid progress in recent years, with methods such as Neural Radiance Fields (NeRF)~\cite{mildenhall2021nerf} and 3D Gaussian Splatting~\cite{kerbl20233d}, along with extensive follow-up work~\cite{dalal2024gaussian}, enabling photorealistic rendering from sparse multi-view input and establishing NVS as a reliable foundation for modern volumetric video pipelines.
The second challenge, relighting, has been extensively studied, but remains difficult in capture stages where inputs are sparse, lighting is uniform, and material or illumination cues are limited.
While one-light-at-a-time (OLAT) capture can enable high-quality relighting~\cite{Xu_2023_ICCV,teufelgera2025HumanOLAT}, it remains impractical for dynamic content, leaving inverse rendering fundamentally underconstrained.
Recent data-driven approaches, including diffusion-based decomposition models, offer a promising alternative by directly predicting material and illumination properties~\cite{ren2024relightful,zhang2025scaling,jin2024neural,liang2025diffusion}, but their applicability in volumetric video pipelines remains unclear.

\input{figures/vram/vram}
This motivates our central question: \emph{Are diffusion relighting models ready for capture-stage compositing?} While their progress is promising, it remains unclear whether existing approaches can meet the temporal stability and scalability demands of real-world volumetric capture pipelines:
We find that diffusion-based relighting models, whose predictions are inherently stochastic, produce inconsistent material predictions under repeated inference, leading to visible flickering when applied frame by frame.
Video-based approaches mitigate this only at the cost of memory requirements that become prohibitive even at moderate resolutions~(cf.~Fig.~\ref{fig:vram}), necessitating inference on short, overlapping frame chunks that introduce temporal discontinuities.

In this work, we revisit relighting from the perspective of capture-stage compositing and explicitly reformulate diffusion-based relighting as a temporally constrained inference problem.
Rather than introducing a new generative diffusion model, we treat state-of-the-art diffusion-based decomposition as a stochastic prior and combine it with explicit temporal regularization and physically motivated rendering.
Concretely, we introduce a flow-guided, confidence-aware temporal regularization scheme, together with a lightweight screen-space relighting module inspired by real-time graphics pipelines.
By combining learned priors with physically informed design choices, our approach outperforms state-of-the-art relighting methods in temporal stability and visual fidelity, making it suitable for volumetric capture pipelines.
In summary, our contributions are
\begin{itemize}
    \item An end-to-end volumetric capture relighting pipeline that integrates state-of-the-art image decomposition priors with physically-motivated rendering, enabling consistent and photorealistic results across dynamic sequences.
    \item A confidence-aware temporal inference formulation that treats stochastic diffusion predictions as noisy observations and aggregates them into temporally coherent material estimates, substantially reducing flickering. %
    \item A comprehensive evaluation of diffusion-based relighting approaches in real-world volumetric capture scenarios, highlighting practical trade-offs in quality, temporal stability, and computation cost.
\end{itemize}

%% file: figures/teaser/teaser.tex
\graphicspath{{figures/teaser/}}

\begin{figure}[t]
  \centering
  \includegraphics[width=\textwidth]{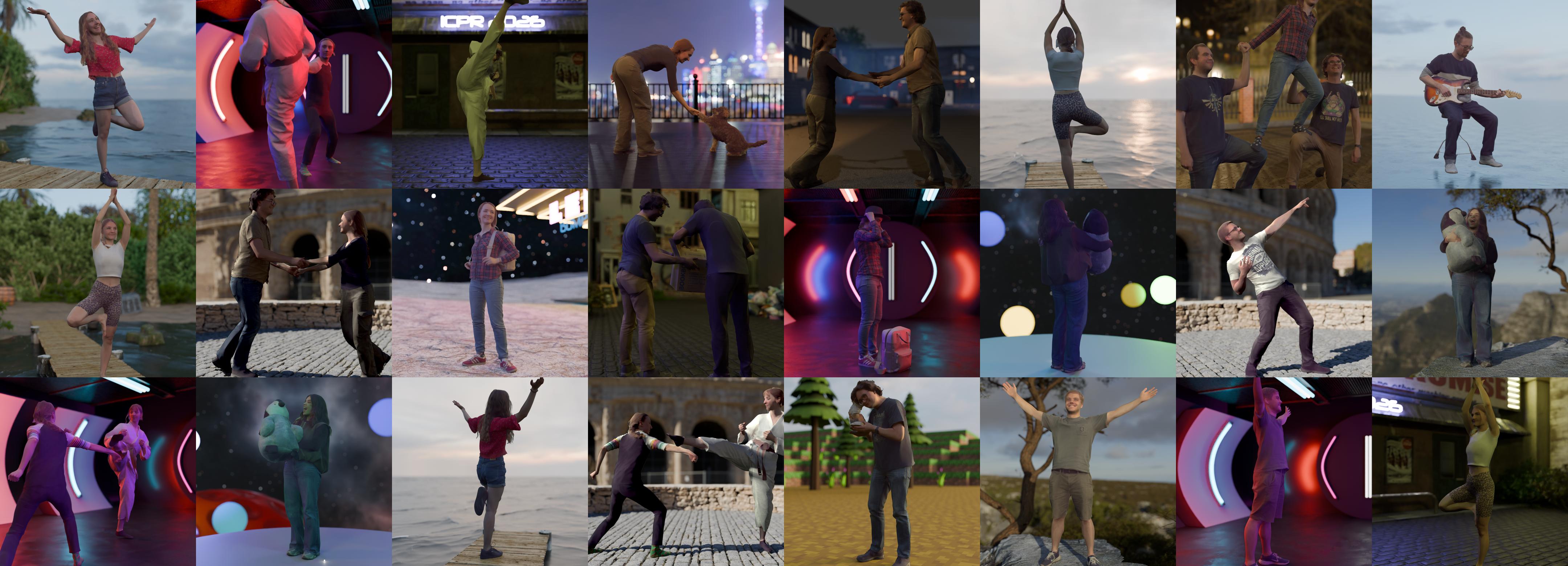}
  \caption{We integrate diffusion-based decomposition priors with variational methods and image-based lighting to achieve physically plausible, temporally stable relighting of captured volumetric content and seamless compositing into virtual environments.}
  \label{fig:teaser}
  \vspace{-14pt}
\end{figure}

%% file: figures/vram/vram.tex
\begin{wrapfigure}[17]{r}{0.4\linewidth}
\vspace{-18pt}
\centering

\includegraphics[width=\linewidth]{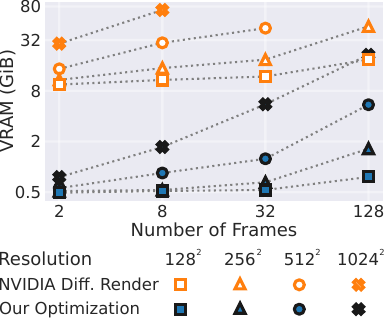}

\caption{
VRAM usage vs.\ resolution and sequence length. The diffusion renderer exceeds memory limits at moderate resolutions even on an 80 GB GPU, while our method scales efficiently to higher resolutions and longer sequences.
}
\label{fig:vram}
\end{wrapfigure}

%% file: sec/2_rel_work.tex
\section{Related Work}
\label{sec:rel_work}

\paragraph{Material Estimation and Relighting}
aim to recover surface reflectance and illumination for appearance synthesis under novel lighting.
Early work formulated this as an ill-posed inverse rendering problem addressed with strong hand-crafted priors \cite{barrow1978recovering, barron2014shape}, including SVBRDF estimation with carefully designed models \cite{aittala2013practical}.
With deep learning, these constraints were relaxed, enabling SVBRDF estimation from single images \cite{deschaintre2018single} and multi-view observations \cite{deschaintre2019flexible}, as well as end-to-end inverse rendering frameworks that jointly recover geometry, reflectance, and illumination using differentiable rendering across single-, dual-, and multi-view settings \cite{yu2019inverserendernet, li2018cgintrinsics, li2020inverse, sengupta2019neural, wimbauer2022rendering, boss2020two, choi2023mair, litman2024materialfusion}.
Recent years have seen significant progress in neural methods for relighting, which aim to synthesize novel illuminations for existing scenes or objects. 
Neural Radiance Fields (NeRF) \cite{mildenhall2021nerf} have been extended to support relighting by explicitly modeling material and illumination properties, via intrinsic decomposition of scene properties \cite{boss2021nerd, zhang2021nerfactor, sarkar2023litnerf, srinivasan2021nerv}, or via a learned scene representation using One-Light-at-a-Time (OLAT) recordings \cite{ Xu_2023_ICCV}.
The introduction of 3D Gaussian Splatting (3DGS) \cite{kerbl20233d} has led to diverse relighting strategies, via intrinsic decomposition \cite{gao2024relightable, du2024gs, liang2024gs, shi2025gir, liang2024gus},  or relighting using OLAT data \cite{dihlmann2024subsurface,bi2024gs3}. %
Some neural relighting methods avoid 3D reconstruction altogether, operating entirely in image space, e.g. by modelling view- and light-dependent transport as a learned function \cite{zhang2021neural} or training a CNN  \cite{xu2018deep} to predict relit results from sparse lighting samples.

More recently, diffusion models have been applied to relighting due to their strong generative priors.
Early works focus on illumination harmonization or background adaptation \cite{ren2024relightful, zhang2025scaling}, while others explore relighting via text prompts or explicit light control \cite{jin2024neural, Zeng_2024, zeng2024dilightnet}.
Neural Gaffer \cite{jin2024neural} supports both text- and environment-map–based relighting.
While these approaches produce convincing results for single images, Diffusion Renderer \cite{liang2025diffusion} extends it to video by unifying inverse and forward rendering, reducing temporal inconsistency at the cost of substantial memory requirements.

\paragraph{Novel-View Synthesis}
aims to generate images of a scene from viewpoints not present in the input observations. 
Early approaches relied on image-based rendering and depth-guided warping, later extended by learning-based approaches using multiplane images or voxel embeddings \cite{eisemann2008floating, goesele2010ambient, chaurasia2011silhouette, Chaurasia:IBR:2013, flynn2019deepview, he2020deepvoxels++, lombardi2019neural}.
More recently, neural radiance fields became the prevalent approach for novel view synthesis,
with subsequent work improving efficiency through alternative scene representations such as low-rank grids and sparse voxels \cite{mildenhall2021nerf, chen2022tensorf, fridovich2022plenoxels}.
NeRFs have been extended to dynamic scenes and relightable representations \cite{pumarola2021d, park2021nerfies, park2021hypernerf, fridovich2023k, boss2021nerd, zhang2021nerfactor, boss2021neural, srinivasan2021nerv}, with further work addressing view-dependent appearance, faces, and scene editing \cite{verbin2022ref, sarkar2023litnerf, yuan2022nerf}.
Real-time rendering has likewise become a key focus \cite{muller2022instant}, and recent work has moved toward alternative representations such as 3D Gaussian Splatting \cite{kerbl20233d}, which replaces implicit fields with explicit point-based primitives for real-time rendering.
Subsequent work has further improved rendering speed and quality \cite{fan2024lightgaussian, girish2024eagles, wang2024sg, liu2024atomgs}
and extended Gaussian-based methods beyond basic rendering, including subsurface scattering and real-time relighting \cite{NEURIPS2024_dc72529d, Saito_2024_CVPR}, as well as surface reconstruction for geometry recovery \cite{yu2024gaussian, PGSR_Chen}.

%% file: sec/3_method.tex
\section{Hybrid Relighting Framework}
\label{sec:method}

\input{figures/overview/overview}

Our method combines three key components as shown in \cref{fig:overview}: volumetric scene representations for novel view synthesis and proxy geometry (\cref{sec:novel_view_synthesis}), estimation and regularization of intrinsic material properties (\cref{sec:gbuffer,sec:temporal_consistency}) and a physically based relighting stage that produces view-consistent shading under novel illumination (\cref{sec:relighting}). The final compositing step accounts for indirect interactions such as cast shadows and reflections, ensuring that the subject integrates plausibly with its surroundings.
The overall pipeline is
designed for offline relighting and compositing and aligns with standard
volumetric capture workflows. In the following, we describe each of these stages in detail.

\subsection{Novel View Synthesis \& Indirect Interactions}
\label{sec:novel_view_synthesis}

Given a multi-camera capture of a performance, our goal is to render the subject from novel viewpoints and to integrate it into a new environment with plausible \textit{indirect {light} interactions}, specifically cast shadows on nearby receivers and reflection contributions, while supporting \textit{relighting}. 
To achieve this, we ideally need a NVS representation that provides (i) RGB appearance, (ii) depth maps, (iii) surface normals, and (iv) a proxy geometry suitable for integration into conventional rendering engines.
Gaussian Opacity Fields (GOF)~\cite{yu2024gaussian} are a natural choice, as they jointly optimize for appearance and normals, produce accurate depth estimates, and allow mesh extraction for downstream use.

However, we observe that GOF reconstructions often suffer from artifacts: curved, natural surfaces are approximated as piecewise planar surfaces, and spurious holes may appear. To mitigate these issues, we introduce two additional regularization terms during optimization: a Huber-regularized total variation loss on the rendered depth map, and an edge-aware Laplacian smoothness prior on the Gaussian normal map.
For the rendered depth map $z(x,y)$, enforce spatial smoothness via 
\begin{equation}
\mathcal{L}_{\text{depth}} = \sum_{x,y} H_\delta(\tfrac{\partial z}{\partial x}) + H_\delta(\tfrac{\partial z}{\partial y}) ,
\end{equation}
where $H_\delta$ denotes Huber loss. %
To prevent blocky surface approximations while respecting appearance discontinuities, we penalize the Laplacian of the estimated normal field $n(x,y)$ in an edge-aware fashion with respect to the RGB image $I(x,y)$ as
\begin{equation}
\mathcal{L}_{\text{normal}} = 
\sum_{x,y} 
\exp\left(- \tfrac{\|\nabla I(x,y)\|^{2}}{2\sigma^{2}} \right)
\cdot \|\Delta n(x,y)\|^{2},
\end{equation}
where $\sigma$ is a contrast sensitivity parameter. Intuitively, smoothing is encouraged in homogeneous regions while being suppressed at edges aligned with appearance boundaries.

\paragraph{Proxy Mesh for Indirect Interactions.}

The extracted GOF mesh serves as a {proxy mesh} that enables indirect light transport effects in conventional rendering engines such as Blender. 
The proxy mesh is not rendered directly; instead, it is included in the scene only to enable indirect interactions such as shadow casting and reflection contributions, without requiring perfect geometry or material accuracy. To this end, intersections with foreground camera rays are disabled, ensuring that the mesh itself remains invisible in the final image.
Since indirect effects are typically low-frequency, coarse geometry suffices for the majority of capture scenarios to render plausible background scenes.

\subsection{Stochastic Intrinsic Decomposition} \label{sec:gbuffer}

We estimate the unknown G-buffer channels (i.e. roughness and metallic)  using learned predictions for view- and light-independent material properties. 
We consider the problem of predicting roughness $r$ and metallic $\mu$ maps, from a single color image $I \in \mathbb{R}^{H \times W \times 3}$ by employing the intrinsic decomposition model $\mathcal{D}_\theta$ from~\cite{liang2025diffusion} that provides the mapping
\begin{equation}
    \mathcal{D}_\theta: I \rightarrow \{r, \mu\}.ö
\end{equation}

Since diffusion sampling is stochastic, we draw $K$ samples per input in order to estimate both mean predictions and per-pixel uncertainty.
To allow high-resolution processing, inference is carried out on overlapping patches $I^{(j)}$ as
\begin{equation}
 \hat{x}^{(i,j)} = \mathcal{D}_\theta(I^{(j)};\,\xi_i), \quad i=1,\dots,K
\end{equation}
where $\xi_i$ denotes the diffusion noise.
These patchwise predictions are then blended into full-resolution maps $\tilde{x}^{(i)}$, yielding an ensemble $\{ \tilde{x}^{(1)},\dots,\tilde{x}^{(K)}\}$
from which we compute per-pixel means and variances, thereby suppressing sampling variability and exposing an uncertainty signal useful for downstream processing.
We treat the captured RGB as a base color approximating diffuse reflectance under near-uniform capture illumination. While not fully intrinsic albedo, it is stable for relighting and compositing and avoids intrinsic decomposition failures.

\subsection{Temporal Consistency via Optical Flow-Guided Regularization}\label{sec:temporal_consistency}
{Next, we address temporally consistent material estimation from stochastic diffusion predictions, treating per-frame diffusion outputs as noisy observations of an underlying time-consistent material field.}
Let $\tilde{x}\in\{\tilde{r},\tilde{\mu}\}$ denote the per-pixel mean prediction and $\sigma_x$ the corresponding normalized standard deviation obtained from multiple diffusion samples. 
We further define a binary foreground mask $m_\alpha = \mathbbm{1}_{\alpha > 0.5}$, backward warping operator $\mathcal{W}$ from frame $t+1$ to $t$ based on optical flow, and a validity mask $\mathcal{V}$ that indicates flow vectors which point to valid foreground and are forward/backward consistent.
Under this formulation, we estimate temporally consistent material maps by solving the following optimization problem over all frames:
\delete{
Let $m_\alpha = \mathbbm{1}_{\alpha>0.5}$ be a binary foreground mask, and let $\mathcal{W}$ denote backward warping from frame $t+1$ to $t$ using estimated optical flow. We write $\odot$ for element-wise multiplication
and $\mathcal{V}$ to denote a binary map of valid flows which point to valid foreground $\mathcal{W}(m_\alpha)$ and are forward/backward consistent.
\newp{Under this formulation, temporally consistent material estimates are recovered by solving a confidence-weighted optimization that balances noisy diffusion observations with spatial and temporal coherence constraints.} \EJ{Irgendwie sowas damit man beim überfliegen checkt was die optimization soll?}
The temporally regularized estimate $x_t$ at step $t$ is obtained by minimizing}
\begin{equation}
    \underset{x}{\arg\min} \ 
\sum_t  \langle H_\delta(\tilde{x}_t - x_t) ,  1 - \sigma_{x,t} \rangle + \lambda_1 TV_\varepsilon(x_t) 
+ \lambda_2
\left\| \left(\mathcal{W}(x_{t+1}) - x_t\right) \odot \mathcal{V} \right\|_2^2 
\end{equation}
where $\odot$ denotes element-wise multiplication, $\langle \cdot,\cdot\rangle$ is the element-wise inner product, $H_\delta$ is the (element-wise) Huber loss,
and $TV_\varepsilon$ the smoothed isotropic total variation for small $\varepsilon$.
The objective consists of three terms that jointly enforce data fidelity, spatial smoothness, and temporal consistency, where the first term  is a confidence-weighted data fidelity loss that aligns $x$ with the diffusion prediction $\tilde{x}$, with deviations weighted according to prediction confidence (i.e., lower weight in regions of high uncertainty). The second term enforces spatial smoothness through total variation regularization, and the third term constrains temporal consistency by penalizing discrepancies between the current roughness or metallic estimate and its backward-warped counterpart.

\subsection{Relighting and Compositing} \label{sec:relighting}

We perform single-image environment relighting using base color, depth, normals, roughness and metallic estimated as described above. Shading is computed using Disney's Principled BRDF~\cite{burley2012physically} using a Cook–Torrance microfacet BRDF with a GGX normal distribution, Smith geometry term, and Schlick Fresnel approximation. 
Illumination is provided by a high-dynamic-range environment map rendered at the approximate actor location in the target scene.
Relighting is performed in screen space: for each
pixel,
the BRDF is integrated over the full environment map and the resulting illumination is modulated
with a soft shadow term computed via view-space ray marching against the
depth buffer. The accumulated occlusion yields a shadow factor in $[0,1]$.
The gamma-corrected, relit foreground is finally composited over the target
background using the alpha channel from the GOF render.

%% file: figures/overview/overview.tex
\begin{figure*}[t]

\centering
\includegraphics[width=\textwidth]{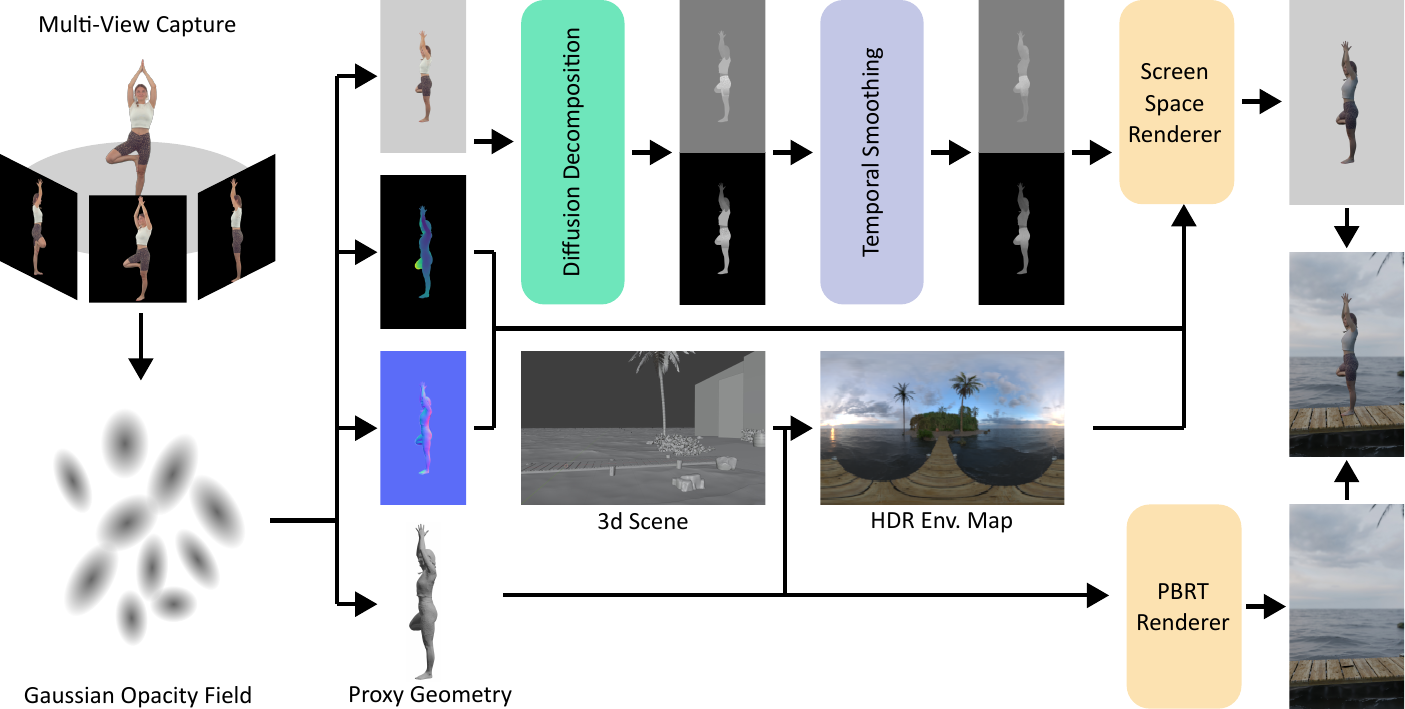}

\caption{We optimize a Gaussian Opacity Field~\cite{yu2024gaussian} from multi-view captures to render RGB, depth and normal maps for the novel views and extract a proxy mesh (left). Using a diffusion decomposition model~\cite{liang2025diffusion} we extract roughness and metallic maps, which we smooth using an optical-flow guided temporal regularization (top). We render the proxy geometry as a shadow caster in the 3d scene (bottom) and blend it with out screen space rendered image (right).}
\label{fig:overview}
\end{figure*}

%% file: sec/4_evaluation.tex
\section{Experiments}
\label{sec:eval}
We benchmark our approach against recent state-of-the-art methods in neural relighting and rendering, including NVIDIA Diffusion Renderer~\cite{liang2025diffusion}, Neural Gaffer~\cite{jin2024neural},  IC-Light~\cite{zhang2025scaling}, and { Relightable 3D Gaussian (R3DG)~\cite{gao2024relightable}.
As reference, we additionally relight our reconstructed proxy mesh using Blender.
}

\paragraph{Datasets and Implementation Details}
 
Our real-world evaluation is based on various scenes from the RIFTCast dataset~\cite{Zingsheim2025RIFTCast}.
We constructed $12$ diverse target environments in Blender to place the captured content.
Furthermore, we created a synthetic dataset of 10 rendered objects in a virtual light stage imitating the above setup. The objects cover diverse geometry and material properties including diffuse and specular reflectance, and are evaluated under 11 HDR environment maps.
All experiments are implemented in PyTorch. We use the Adam optimizer %
with its default parameters.

\subsection{Synthetic Analysis of Relighting Accuracy} \label{sec:quantitativ}
\input{table/quantitative}

We quantitatively evaluate relighting accuracy on the synthetic dataset by comparing relit outputs against ground-truth renderings under identical environment maps. \Cref{tab:quantitative} summarizes the results across standard image-based metrics.
Our method consistently achieves the best scores across all three metrics. In particular, it improves PSNR by a clear margin over all competing diffusion-based approaches, while also yielding the highest SSIM and lowest LPIPS.
\delete{
This is consistent with the qualitative comparison, where certain methods generated attractive images but failed to capture the true characteristics of the lighting in the scene that ours preserves.
}
\delete{
\HD{Das muss noch raus/angepasst werden:}
GOF serves as a strong geometric reference, yet it is limited by hard surface boundaries and struggles in regions with complex transparency, as already seen qualitatively. In contrast, our pipeline maintains both high fidelity and perceptual quality across diverse lighting conditions.}
\cref{fig:synthetic_qualitative} further provides qualitative comparisons on the synthetic dataset, showing our relit results alongside ground-truth renderings.

\paragraph{Memory Requirements}

As shown in \cref{fig:vram}, the memory consumption of the Diffusion Renderer -- the second-best performing method -- increases steeply with both spatial resolution and sequence length. This requires inference on short, overlapping frame chunks, which limits scalability and introduces temporal discontinuities. In contrast, our method supports high-resolution processing with significantly lower and more predictable memory requirements, enabling temporally consistent relighting across long sequences.
The Diffusion Renderer was evaluated at a reduced resolution of 896$\times$512, as inference at full resolution (1920$\times$1080) additionally  resulted in severe artifacts and degraded  quality.

\input{figures/qualitative2_temporal}

\input{figures/qualitative/qualitative}
\input{figures/alpha/alpha}
\subsection{Real World Application}\label{sec:qualitative}

Qualitative evaluation on real-world data (see \cref{fig:qualitative}) indicates that prior methods face challenges in faithfully reproducing scene appearance, either by failing to model correct illumination, as in Neural Gaffer and DiffRenderer, or even by suffering from noticeable color inaccuracies, as in R3DG and IC-Light.
In contrast, we observe that our method produces more faithful illumination and, relative to the relighting using a proxy mesh in Blender, yields smoother surface reconstructions, as illustrated by the arm shown in \cref{fig:alpha}, together with more realistic alpha mattes in challenging regions such as hair.
The resulting increase in visual realism is further supported by a user study, reported in the supplementary material.
Videos generated using our method are available on the \href{https://ejuet.github.io/yesnt/}{project page}.

\paragraph{Frame-to-Frame Coherence}
Beyond static relighting quality, volumetric capture-stage compositing requires consistency across a sequence.
We observe that our method maintains coherence and preserves stable material appearance, whereas diffusion-only approaches frequently exhibit temporal inconsistency, particularly when constrained to short windows imposed by memory limitations~(see also \cref{fig:temporal_consistency}). These observations are supported by a user study (see supplementary material),
in which participants rated our results as significantly 
more temporally stable,
with a very large effect size for temporal smoothness (Cohen’s $d = 1.57$).

\subsection{Ablation Study}\label{sec:ablation}

\begin{wraptable}[10]{r}{0.55\linewidth}
\vspace{-33pt}
    \centering
    \caption{
    Ablation study of our model, comparing variants without any relighting (\ie novel view synthesis only), without estimation of G-Buffer maps (roughness and metallic), and the full model.}
    \vspace{5pt}
    \setlength{\tabcolsep}{4.0pt}
    \begin{tabular}{l|rrr}
        \toprule
        Variant & PSNR $\uparrow$ & SSIM $\uparrow$ & LPIPS $\downarrow$ \\
        \midrule
        No Rel. & $22.8414$ & $0.9265$ & $0.0742$ \\
        No Decomp. & $28.5688$ & $0.9507$ & $0.0565$ \\
        Full  &$30.1918 $ & $0.9580 $  &  $0.0523$ \\
        \bottomrule
    \end{tabular}
    \label{tab:temporal_ablation}
\end{wraptable}

We conduct an ablation study to assess the contribution of individual components of our model. 
\cref{tab:temporal_ablation} reports results when removing either the relighting branch or the estimation of G-Buffer maps (roughness/metallic). 
Without relighting, the network reduces to a pure NVS model and achieves substantially lower accuracy.
Removing the intrinsic decomposition
and keeping roughness and metallic constant also results in a notable drop across all metrics, showing that physically meaningful intermediate representations are crucial for high-fidelity relighting.
The full model achieves the highest scores across all metrics.

\begin{wraptable}[12]{r}{0.40\linewidth}
\vspace{-33pt}
    \centering
    \caption{
    Temporal regularization ablation on videos (see \cref{sec:temporal_consistency}). Removing temporal regularization (w/o Temporal Reg.) leads to a decrease in temporal PSNR (tPSNR) and larger PPL, indicating stronger flickering artifacts.}
    \vspace{5pt}
    \setlength{\tabcolsep}{4.0pt}
    \begin{tabular}{l|rr}
        \toprule
        Variant & tPSNR $\uparrow$ & PPL $\downarrow$  \\
        \midrule
        w/o Reg.  &$28.8737$ & $0.3598$  \\
        Full  &$29.3660$ & $0.3587$   \\
        \bottomrule
    \end{tabular}
    \label{tab:temporal_ablation_flickering}
\end{wraptable}

We further evaluate the effect of temporal regularization, as introduced in \cref{sec:temporal_consistency}. 
\cref{tab:temporal_ablation_flickering} compares the full model against a variant without temporal regularization. 
The latter suffers from degraded temporal PSNR (tPSNR) and a higher perceptual path length (PPL), indicating the presence of flickering artifacts across video frames. 
Note, that while PPL is typically defined in the context of latent interpolations in generative models, we compute it directly across consecutive output frames to measure frame-to-frame perceptual variation.
\cref{fig:temporal_consistency} illustrates this flickering effect on predicted roughness maps: the Diffusion Renderer~\cite{liang2025diffusion} exhibits strong temporal inconsistency across consecutive frames, whereas our flow-guided smoothing produces temporally coherent results while preserving spatial detail. 
These findings underline the importance of temporal regularization for stable relighting in dynamic scenes.

\newpage
\subsection{Limitations}
\label{sec:limitations}

\input{table/runtimes}

The runtime of our method scales proportionally with sequence length, with the diffusion-based decomposition forming the primary computational bottleneck and limiting throughput for long sequences (cf.~\cref{tab:runtimes}). While this
currently restricts the method to offline processing, it remains compatible
with typical volumetric capture  workflows
as its computational overhead is negligible compared to GOF’s optimization cost.

Furthermore, our method assumes a single environment‐map illumination and does not model spatially varying light sources, which limits local light transport effects (e.g., missing cast shadows on limbs). 

Finally, our method relies on the assumption that diffusion-based material estimates are accurate on average. 
While this holds in most cases, rare failures occur when the diffusion model produces strongly inconsistent predictions across frames. In these cases, temporal regularization may suppress such outliers, leading to over-smoothing of fine-scale detail.

%% file: table/quantitative.tex
\definecolor{best}{rgb}{1, 0.7, 0.7}
\definecolor{second}{rgb}{1, 0.85, 0.7}

\begin{table*}[t]
    \centering
    \caption{Quantitative comparison of relighting accuracy on synthetic data.}%
    \label{tab:quantitative}
    \resizebox{\linewidth}{!}{
    \begin{tabular}{lcccc}
    \toprule
    Method & Conditioning & PSNR $\uparrow$ & SSIM $\uparrow$ & LPIPS $\downarrow$ \\
    \midrule
    Reference
    & HDR Env.\ Map
    & $29.42 \pm 5.79$
    & $0.946 \pm 0.040$
    & $0.062 \pm 0.055$ \\
    \midrule
    IC-Light~\cite{zhang2025scaling}
    & LDR Background
    & $19.41 \pm 4.58$
    & $0.908 \pm 0.081$
    & $0.087 \pm 0.084$ \\
    Neural Gaffer~\cite{jin2024neural}
    & HDR+LDR Env.\ Map
    & $24.43 \pm 4.50$
    & $0.917 \pm 0.084$
    & $0.084 \pm 0.092$ \\
    R3DG \cite{gao2024relightable}
    & HDR Env.\ Map
    & $26.15 \pm 5.92$
    & $0.926 \pm 0.067$
    & \colorbox{second}{$0.078 \pm 0.071$} \\
    Diffusion Renderer~\cite{liang2025diffusion}
    & LDR+Log Env.\ Map
    & \colorbox{second}{$27.49 \pm 6.19$}
    & \colorbox{second}{$0.937 \pm 0.050$}
    & $0.082 \pm 0.071$ \\
    Yesnt (Ours)
    & HDR Env.\ Map
    & \colorbox{best}{$30.19 \pm 5.48$}
    & \colorbox{best}{$0.958 \pm 0.028$}
    & \colorbox{best}{$0.052 \pm 0.046$} \\
    \bottomrule
    \end{tabular}
    }
\end{table*}

%% file: figures/qualitative2_temporal.tex
\begin{figure}[t]
\centering

\begin{minipage}[t]{0.42\linewidth}
\centering

\begin{minipage}[b]{0.31\linewidth}
    \includegraphics[width=\linewidth]{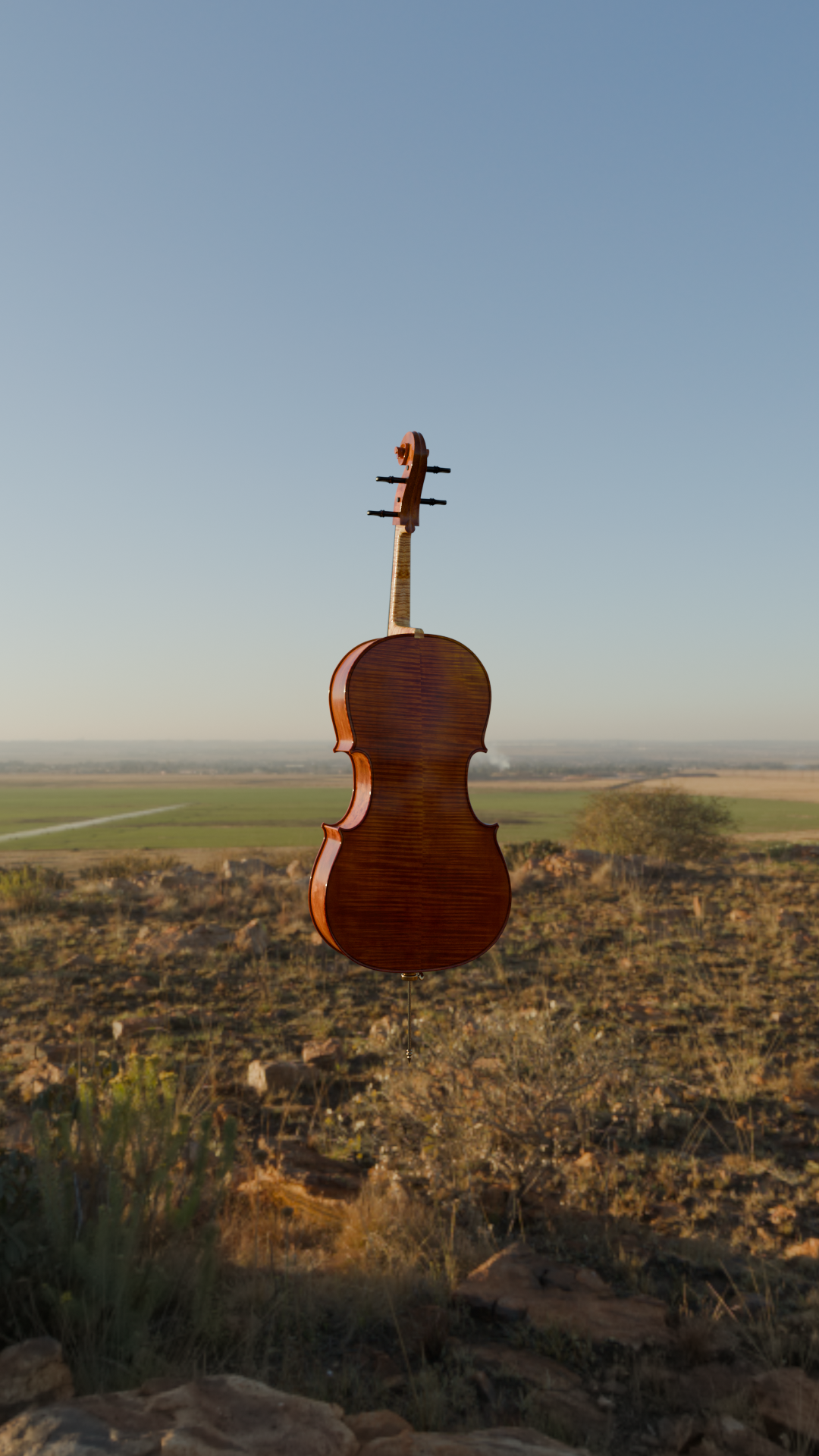}
\end{minipage}
\begin{minipage}[b]{0.31\linewidth}
    \includegraphics[width=\linewidth]{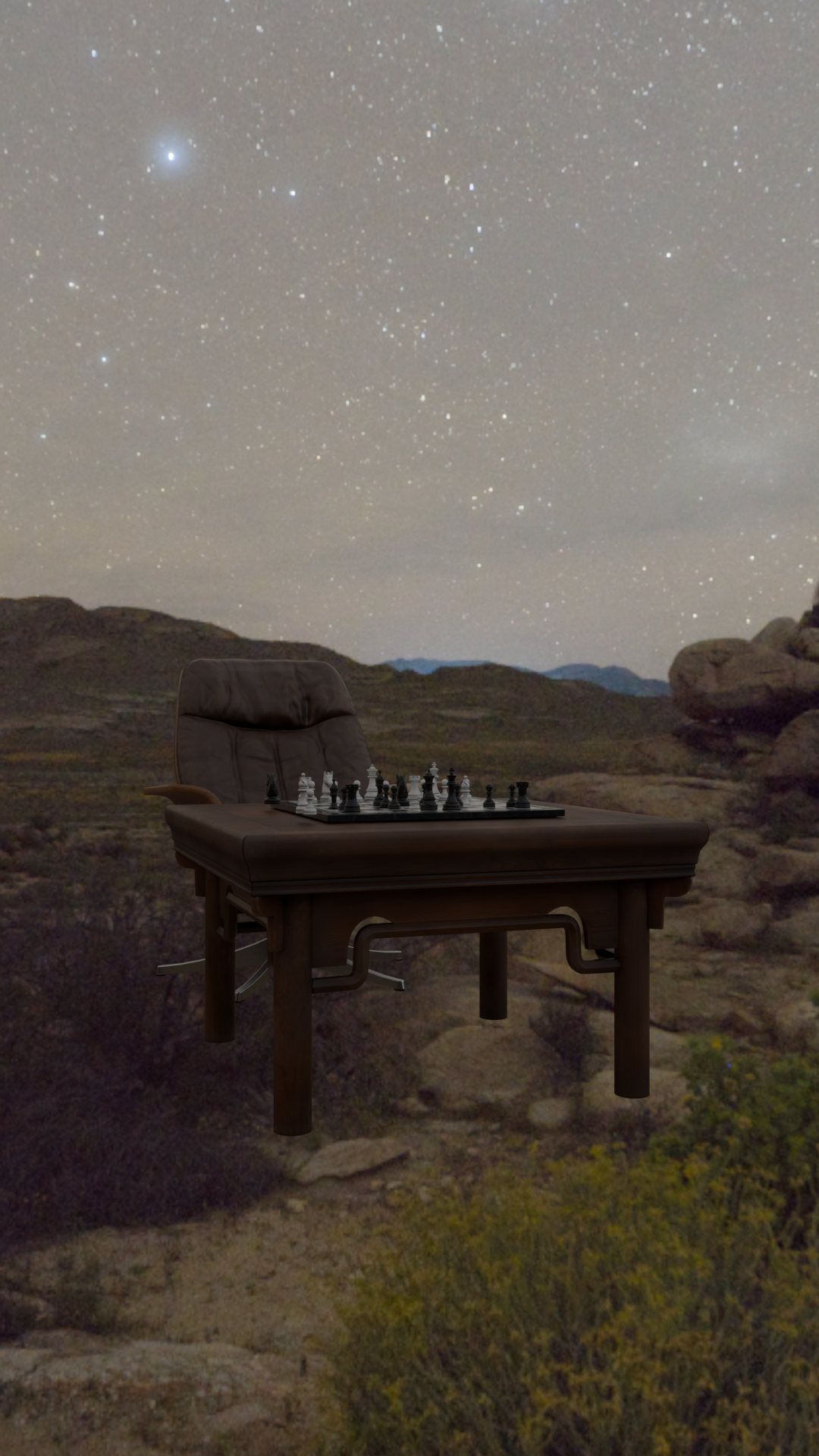}
\end{minipage}
\begin{minipage}[b]{0.31\linewidth}
    \includegraphics[width=\linewidth]{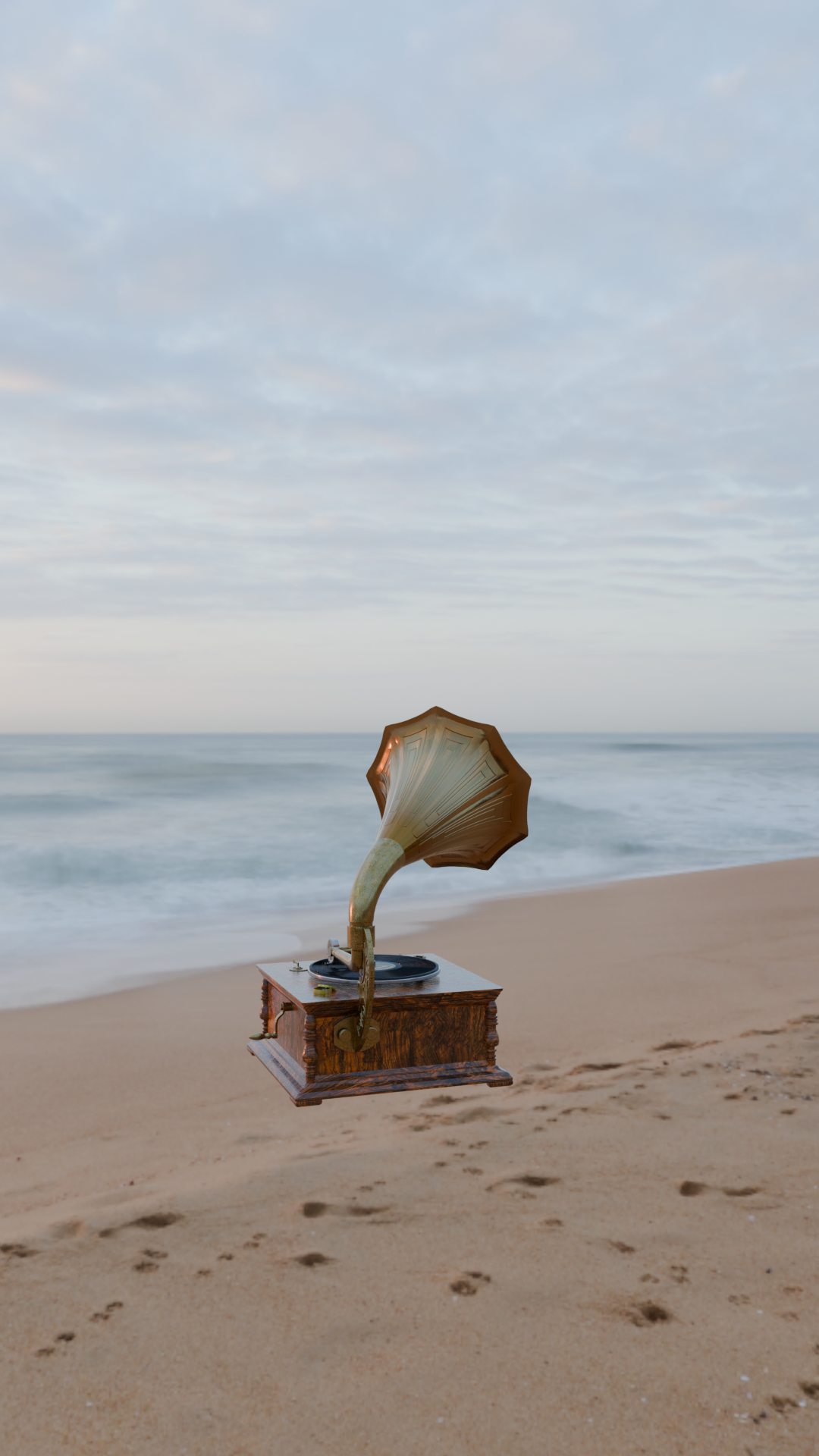}
\end{minipage}

\vspace{0.5em}

\begin{minipage}[b]{0.31\linewidth}
    \includegraphics[width=\linewidth]{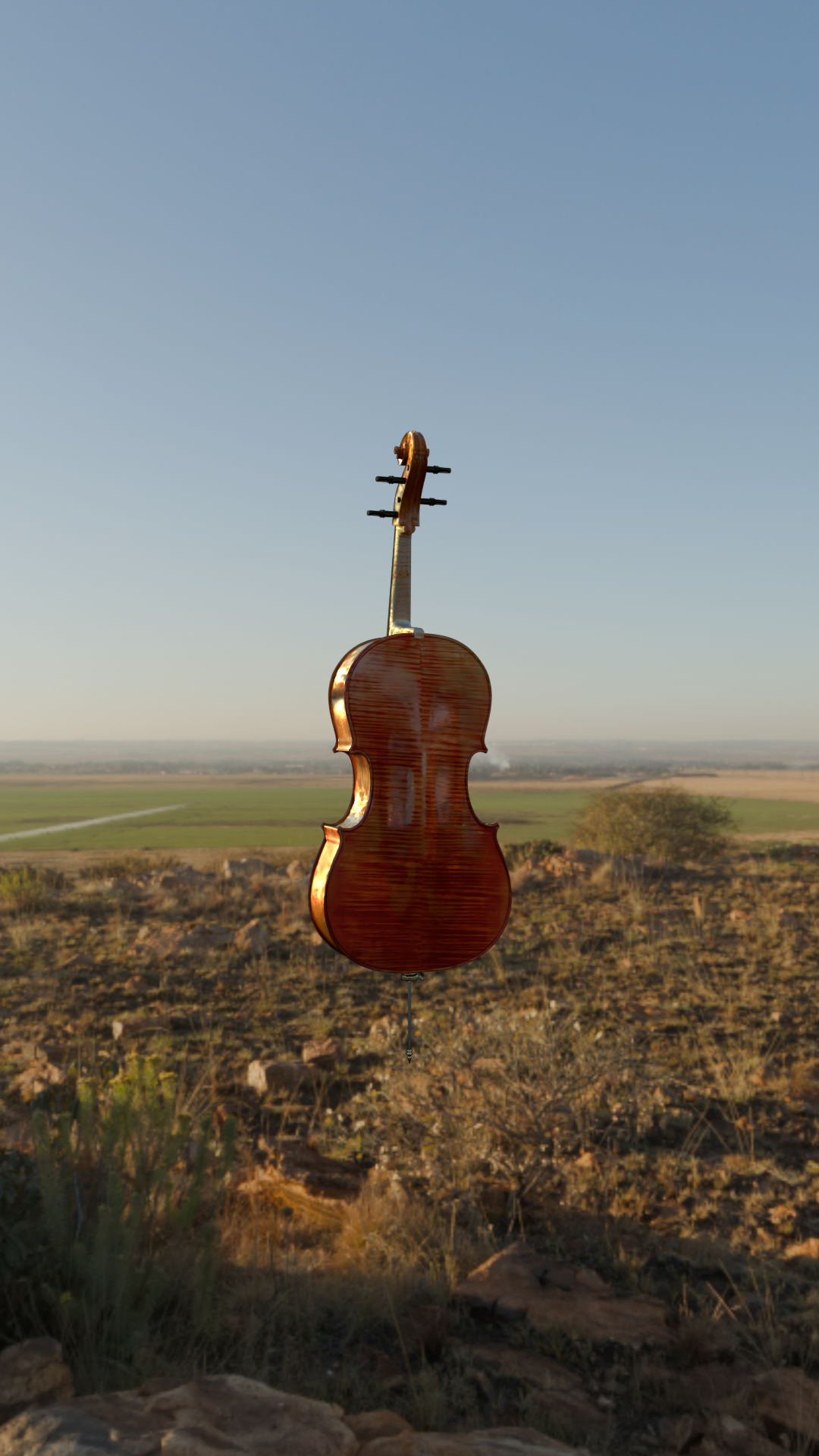}
\end{minipage}
\begin{minipage}[b]{0.31\linewidth}
    \includegraphics[width=\linewidth]{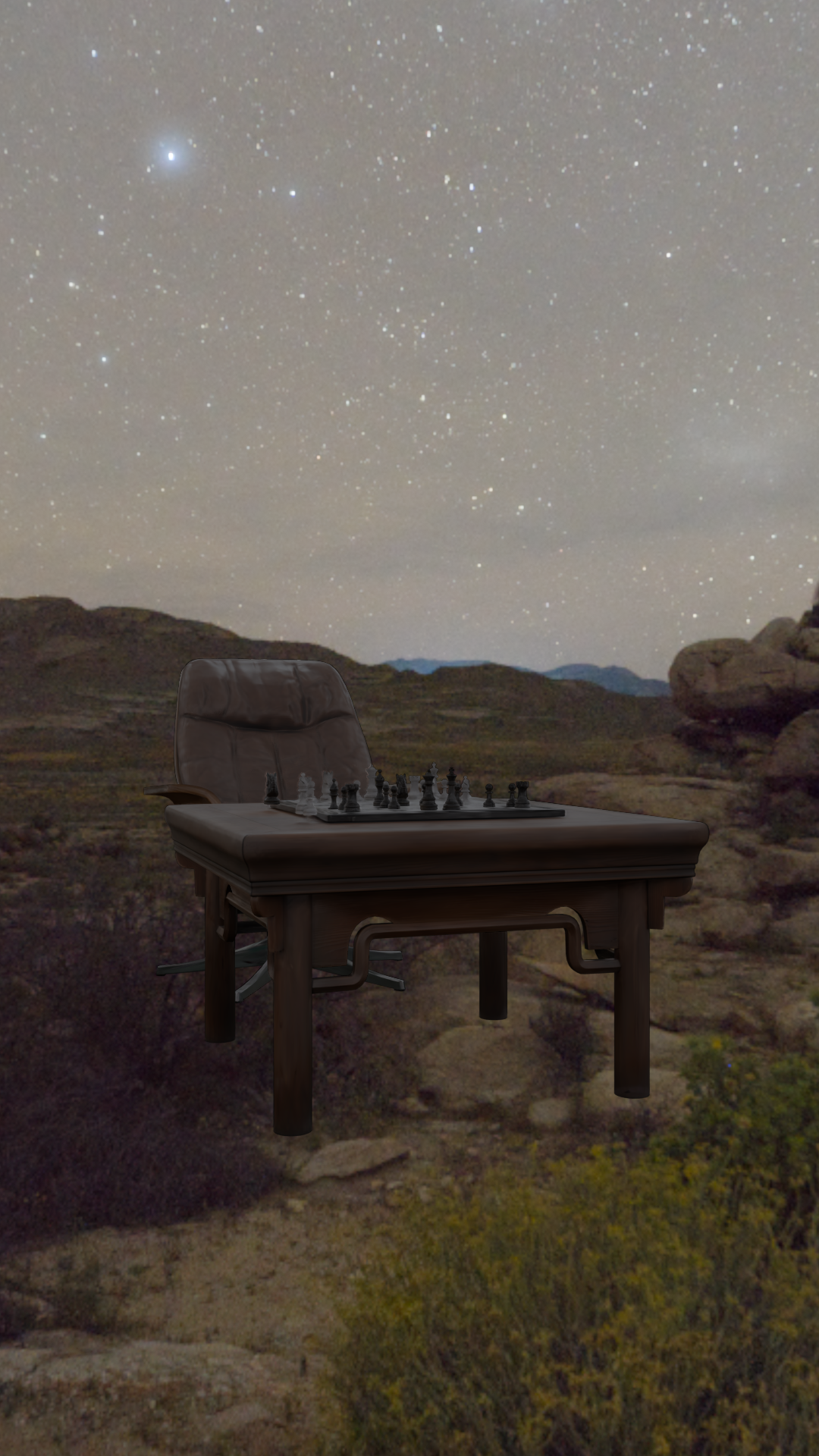}
\end{minipage}
\begin{minipage}[b]{0.31\linewidth}
    \includegraphics[width=\linewidth]{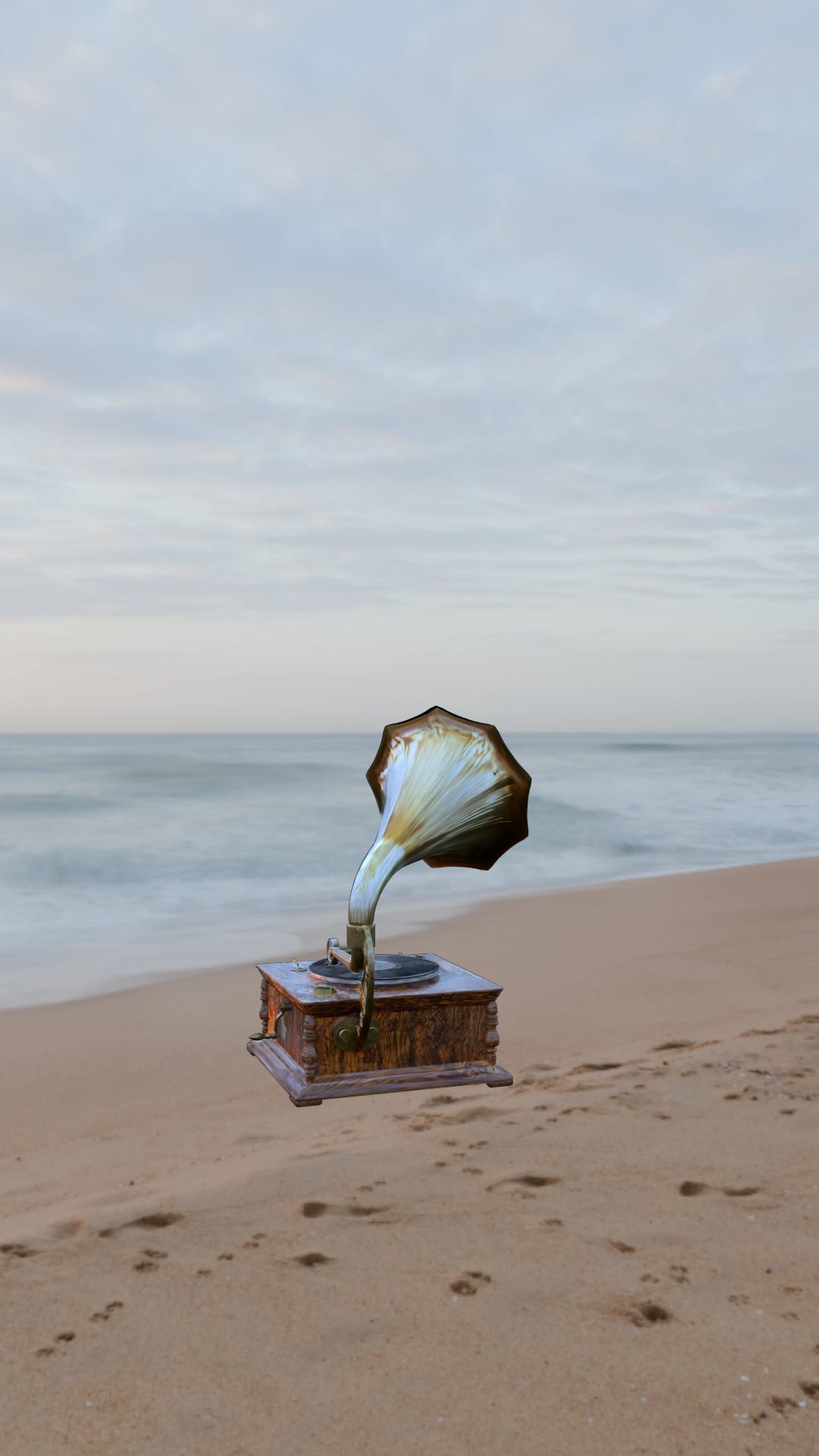}
\end{minipage}

\vspace{-8pt}
\captionof{figure}{Qualitative relighting on the synthetic dataset. Ground truth (top) and our method (bottom).}
\label{fig:synthetic_qualitative}
\end{minipage}
\hfill
\begin{minipage}[t]{0.5375\linewidth}
\centering

\begin{minipage}[b]{0.31\linewidth}
    \includegraphics[width=\linewidth]{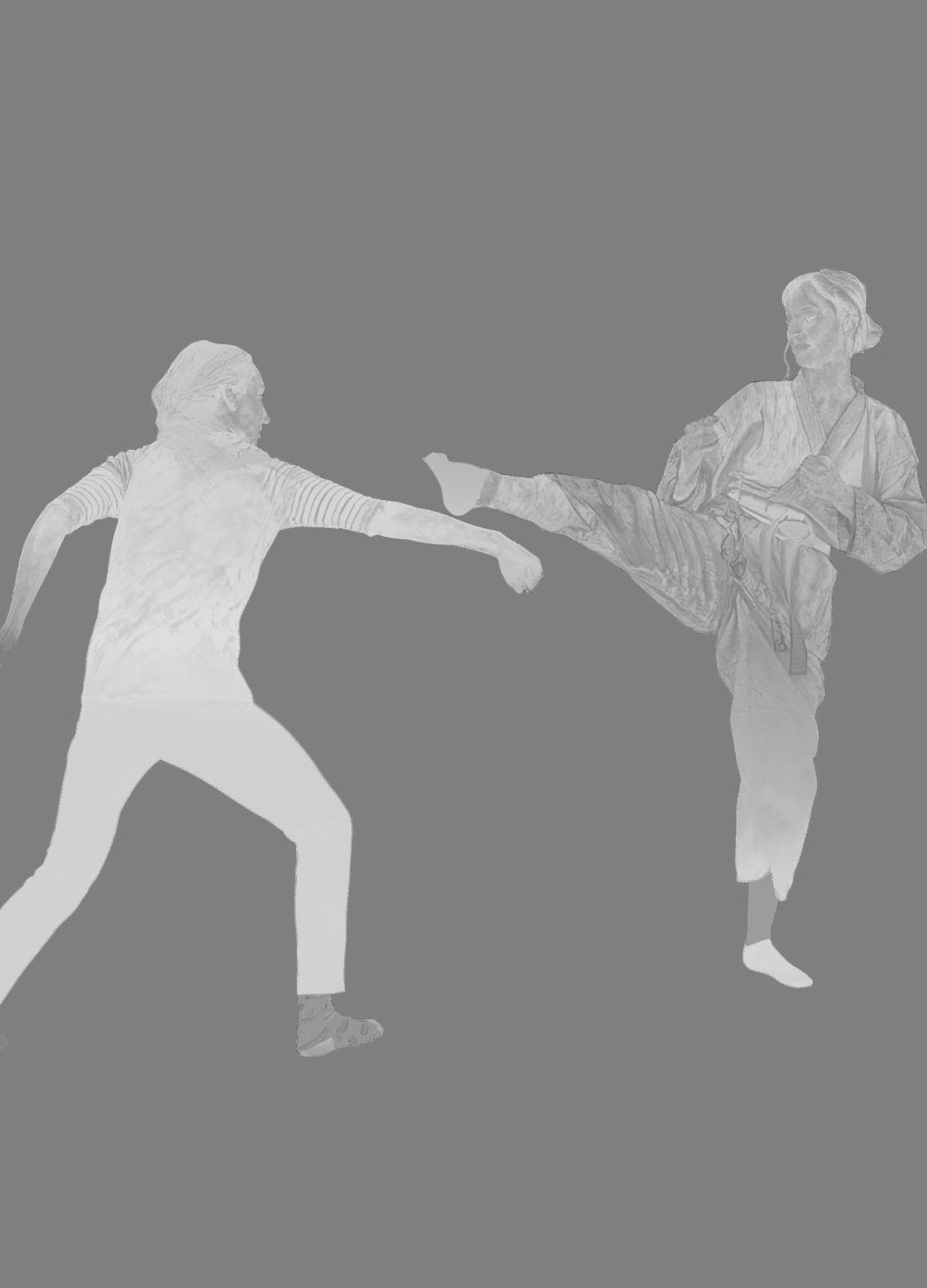}
\end{minipage}
\begin{minipage}[b]{0.31\linewidth}
    \includegraphics[width=\linewidth]{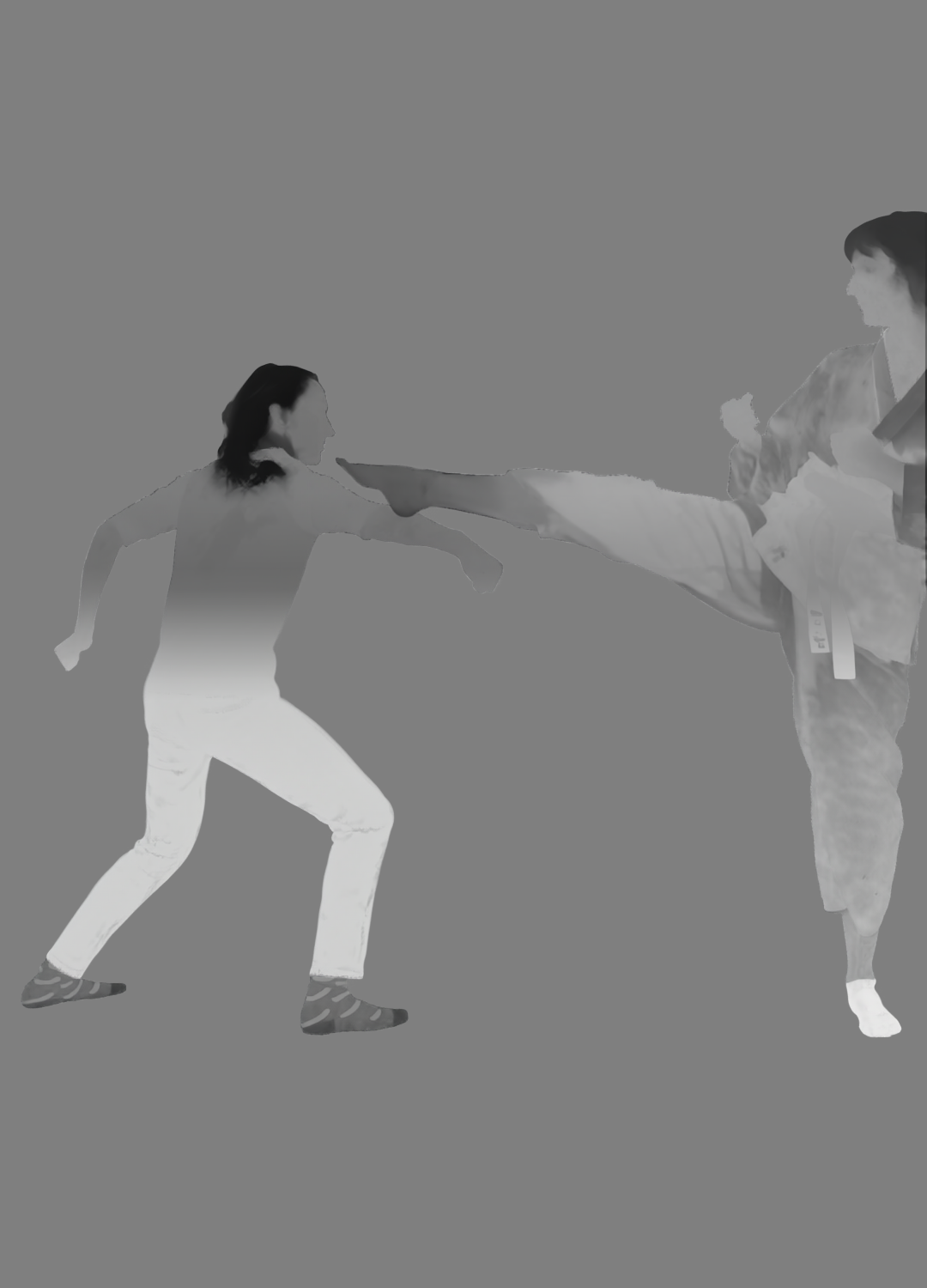}
\end{minipage}
\begin{minipage}[b]{0.31\linewidth}
    \includegraphics[width=\linewidth]{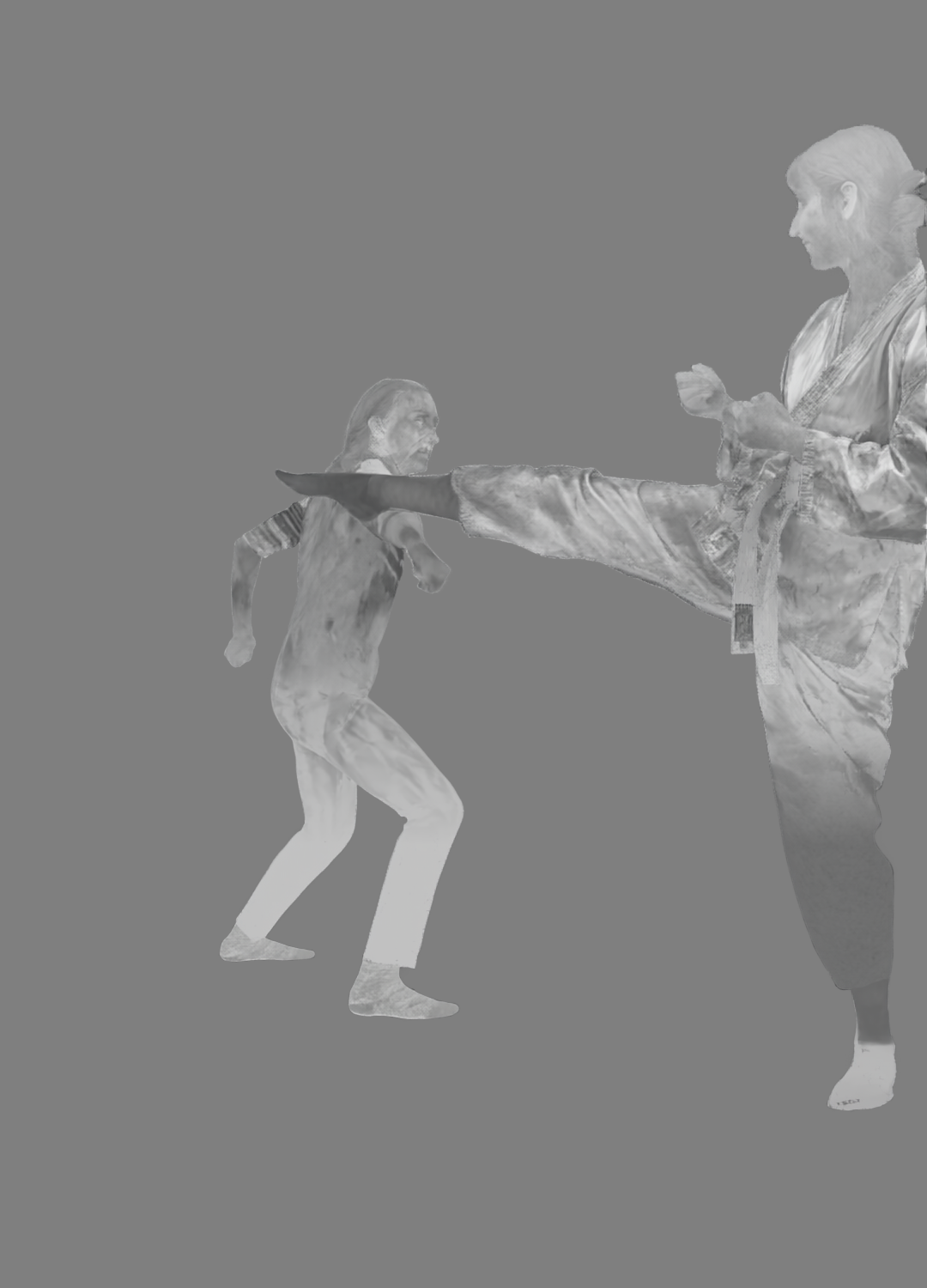}
\end{minipage}

\vspace{0.5em}

\begin{minipage}[b]{0.31\linewidth}
    \includegraphics[width=\linewidth]{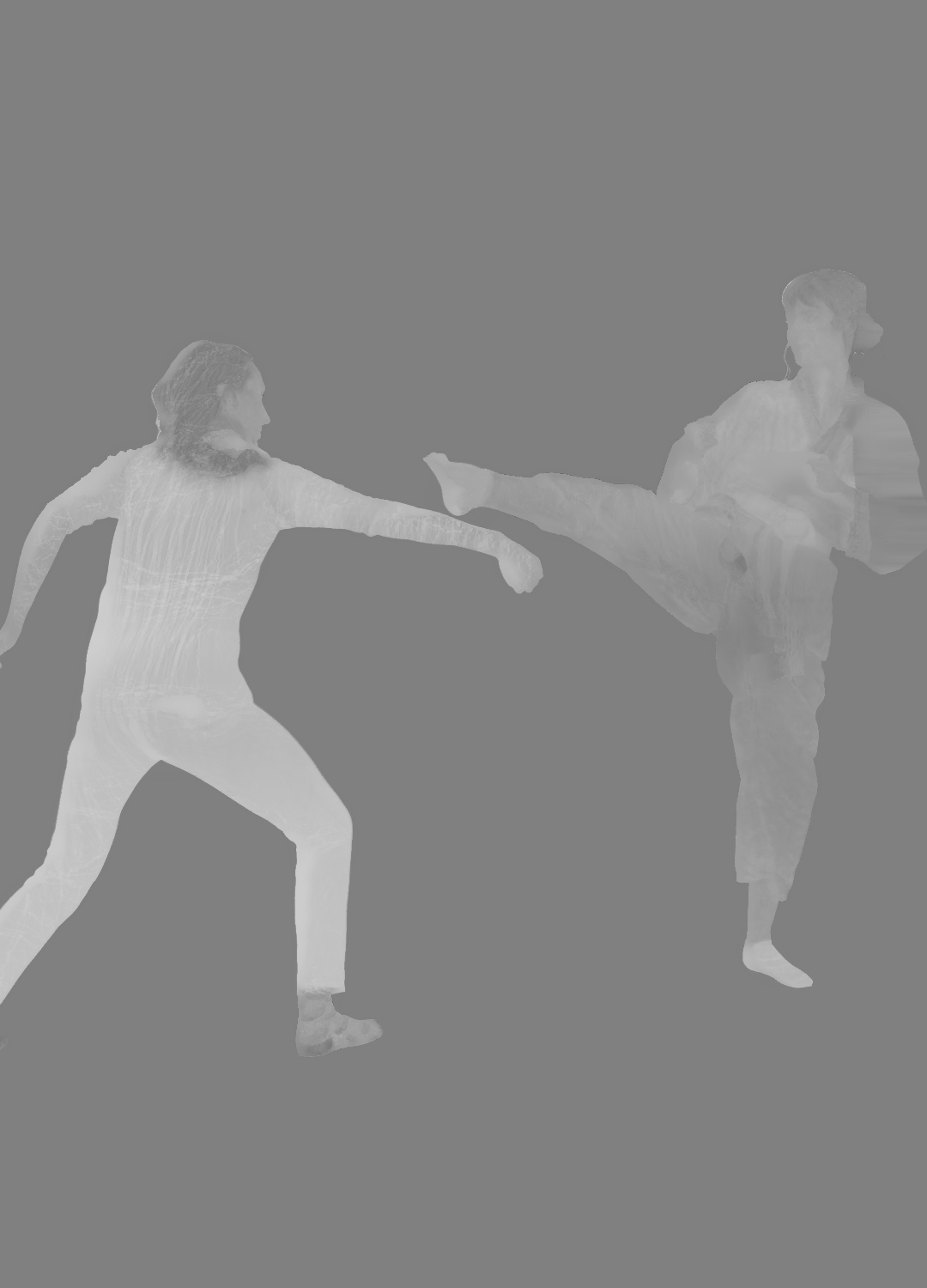}
\end{minipage}
\begin{minipage}[b]{0.31\linewidth}
    \includegraphics[width=\linewidth]{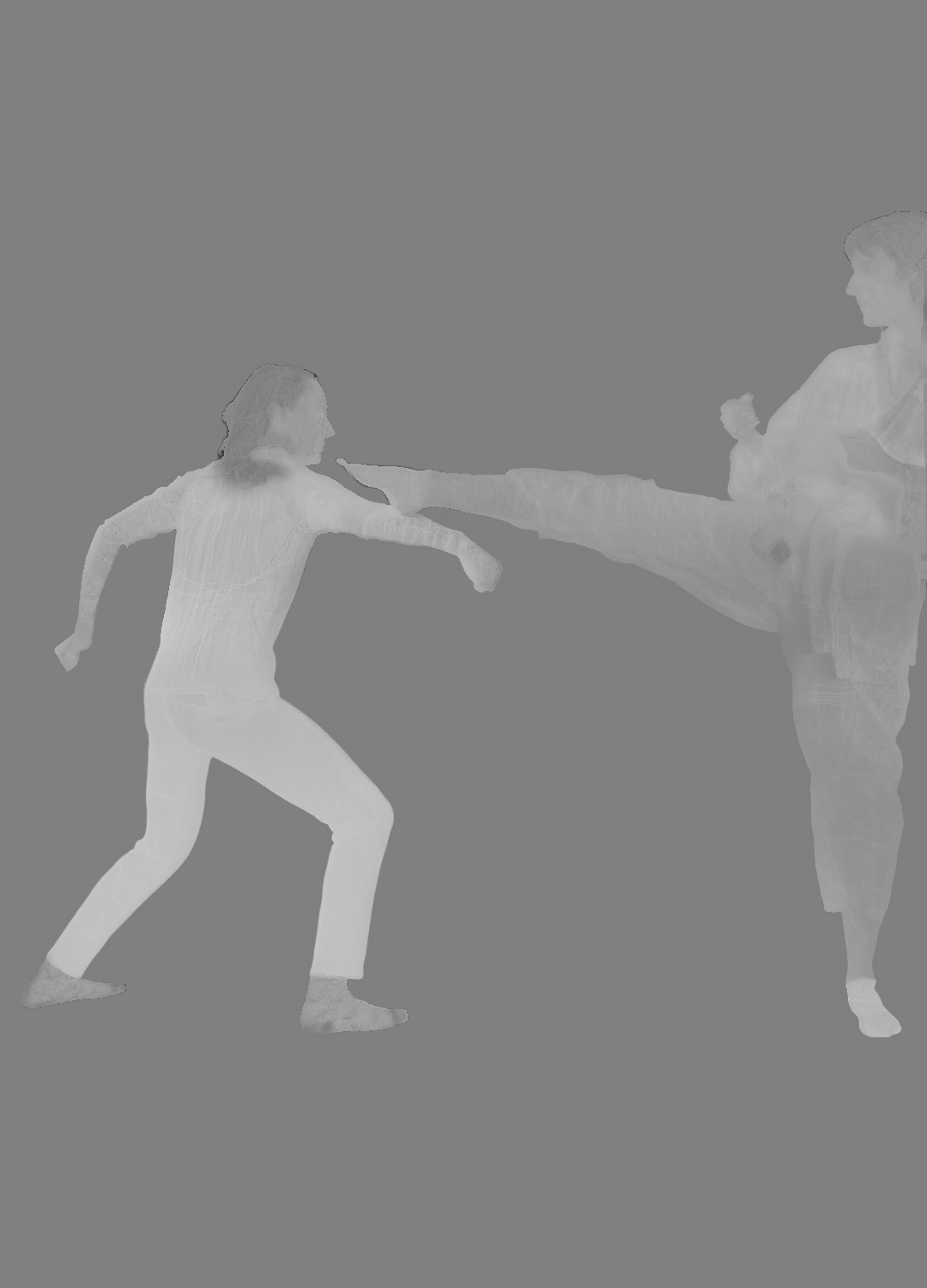}
\end{minipage}
\begin{minipage}[b]{0.31\linewidth}
    \includegraphics[width=\linewidth]{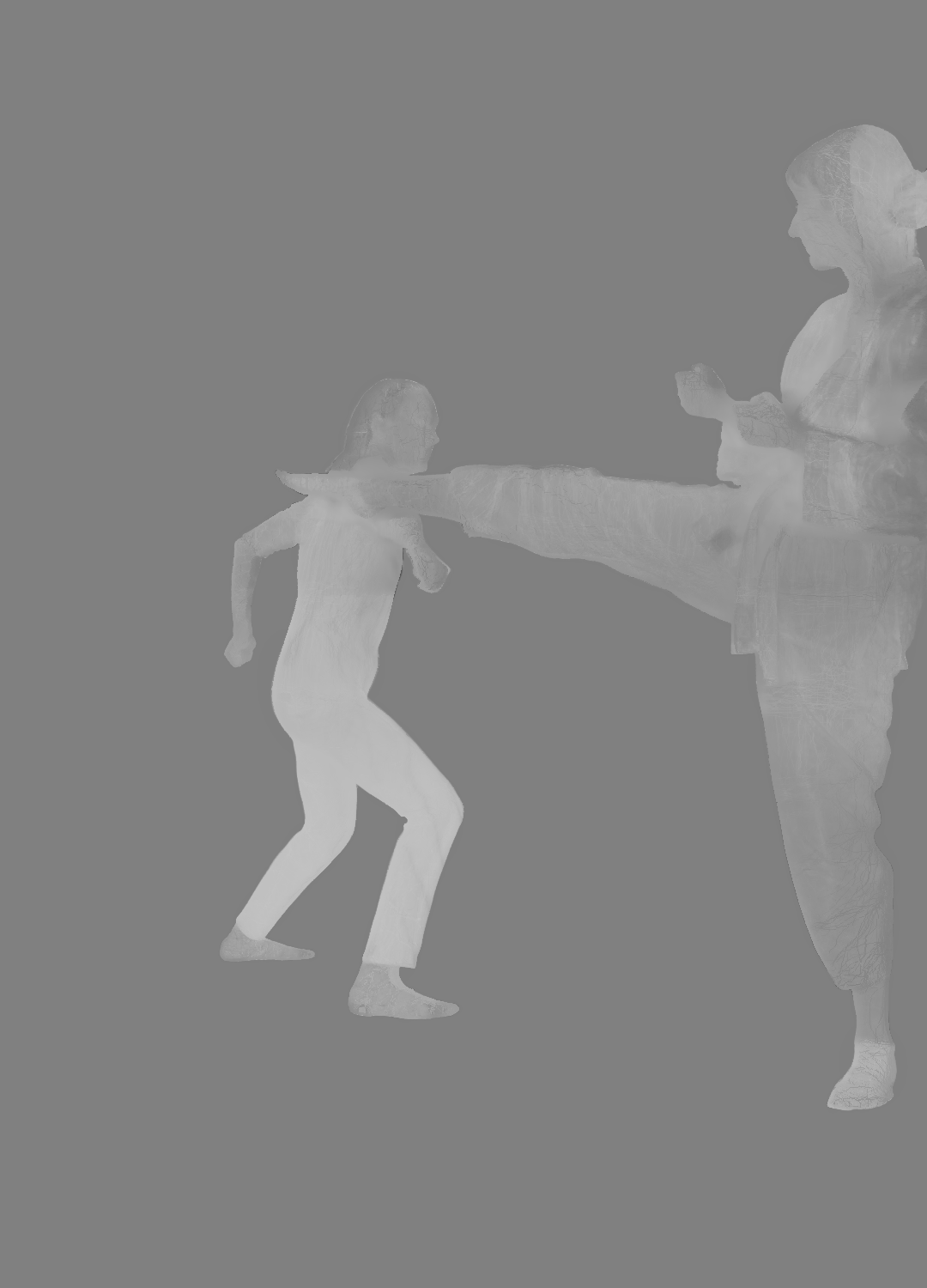}
\end{minipage}

\vspace{-8pt}
\captionof{figure}{
Roughness estimates show flickering in the Diff.\ Renderer (top) and improved temporal stability with our optimization (bottom).
}
\label{fig:temporal_consistency}
\end{minipage}

\end{figure}

%% file: figures/qualitative/qualitative.tex
\begin{figure*}[htb!]

\centering

\begin{minipage}[b]{0.16\linewidth}
    \includegraphics[width=\linewidth]{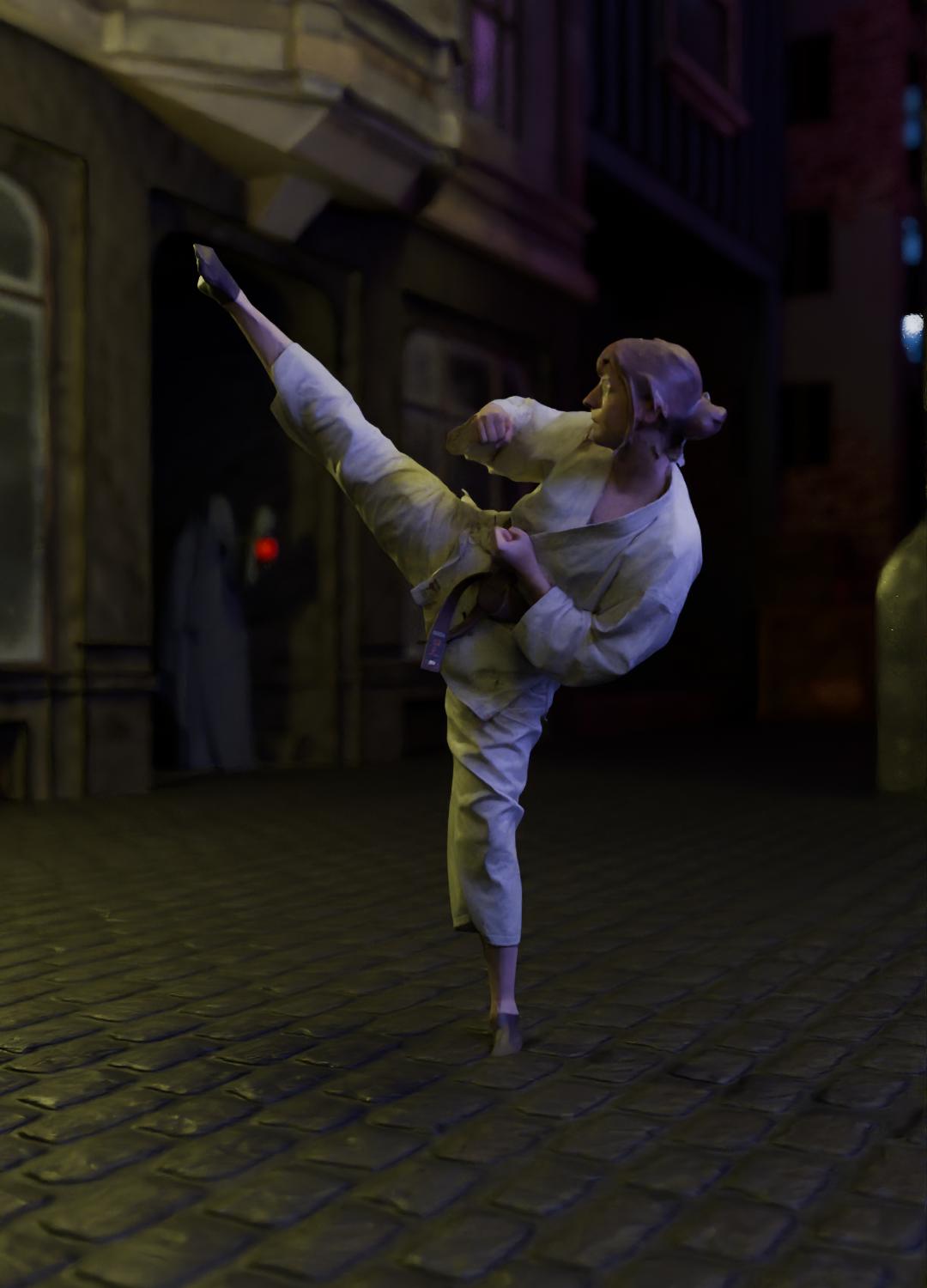}
\end{minipage}
\begin{minipage}[b]{0.16\linewidth}
    \includegraphics[width=\linewidth]{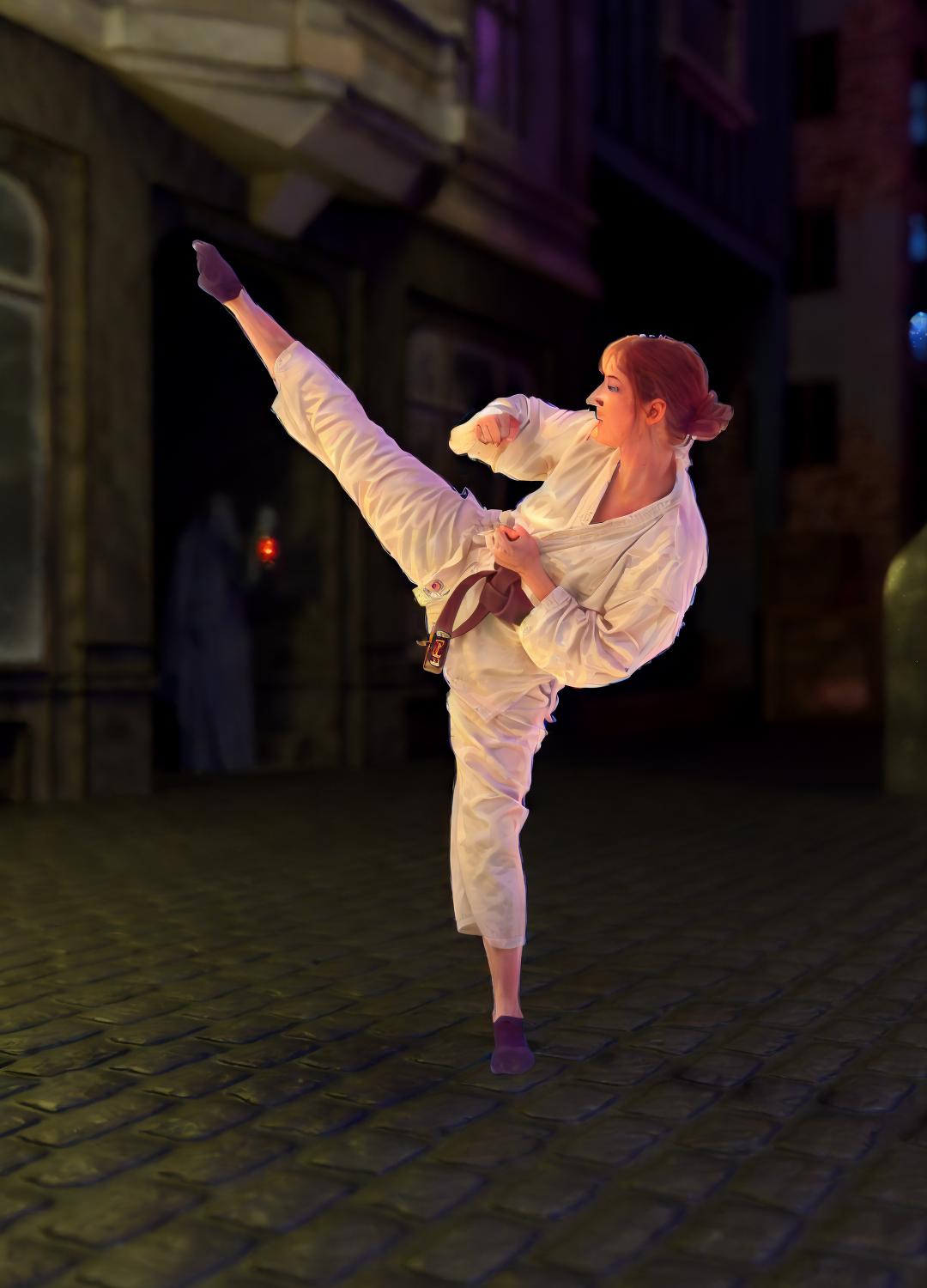}
\end{minipage}
\begin{minipage}[b]{0.16\linewidth}
    \includegraphics[width=\linewidth]{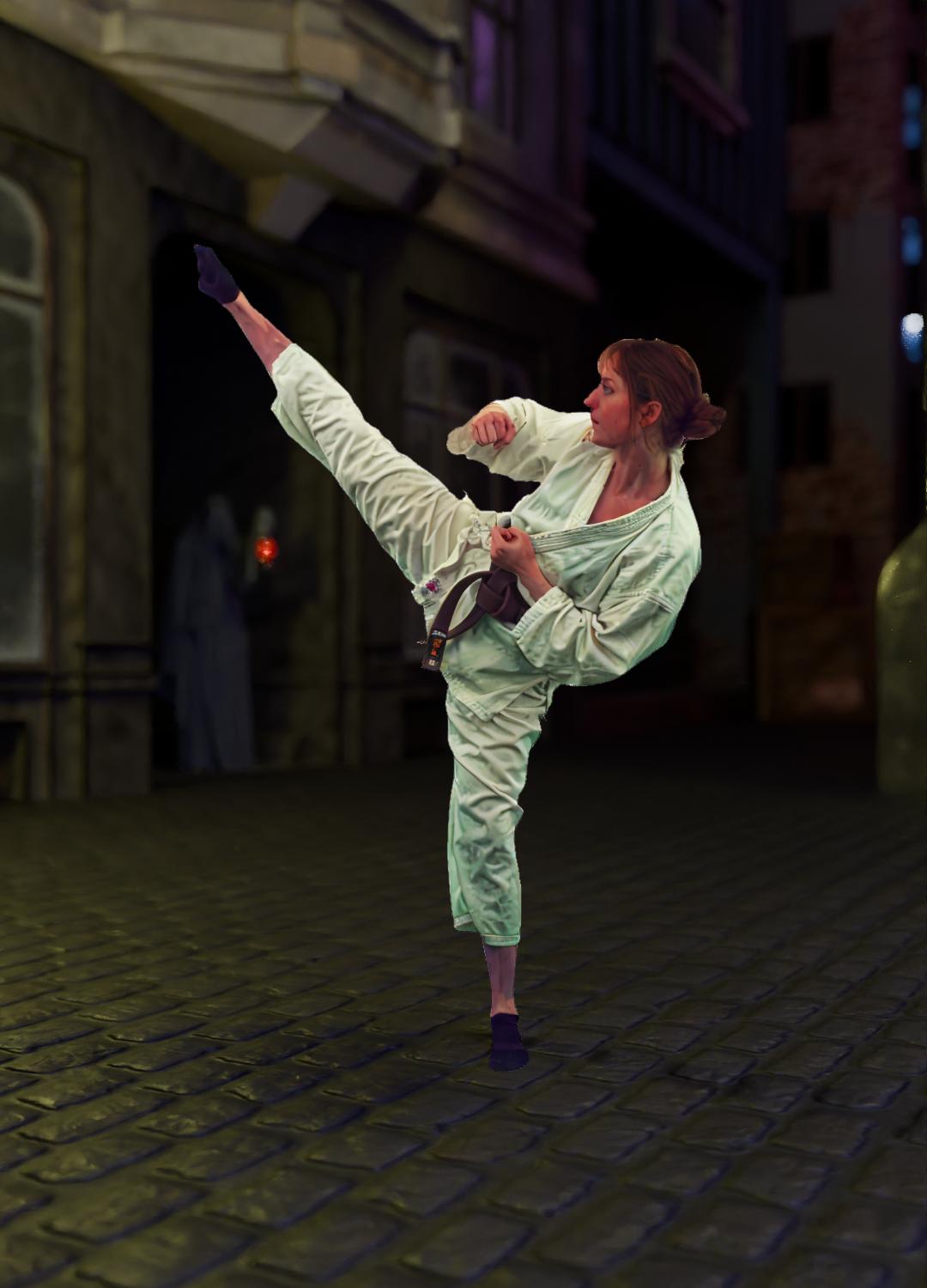}
\end{minipage}
\begin{minipage}[b]{0.16\linewidth}
    \includegraphics[width=\linewidth]{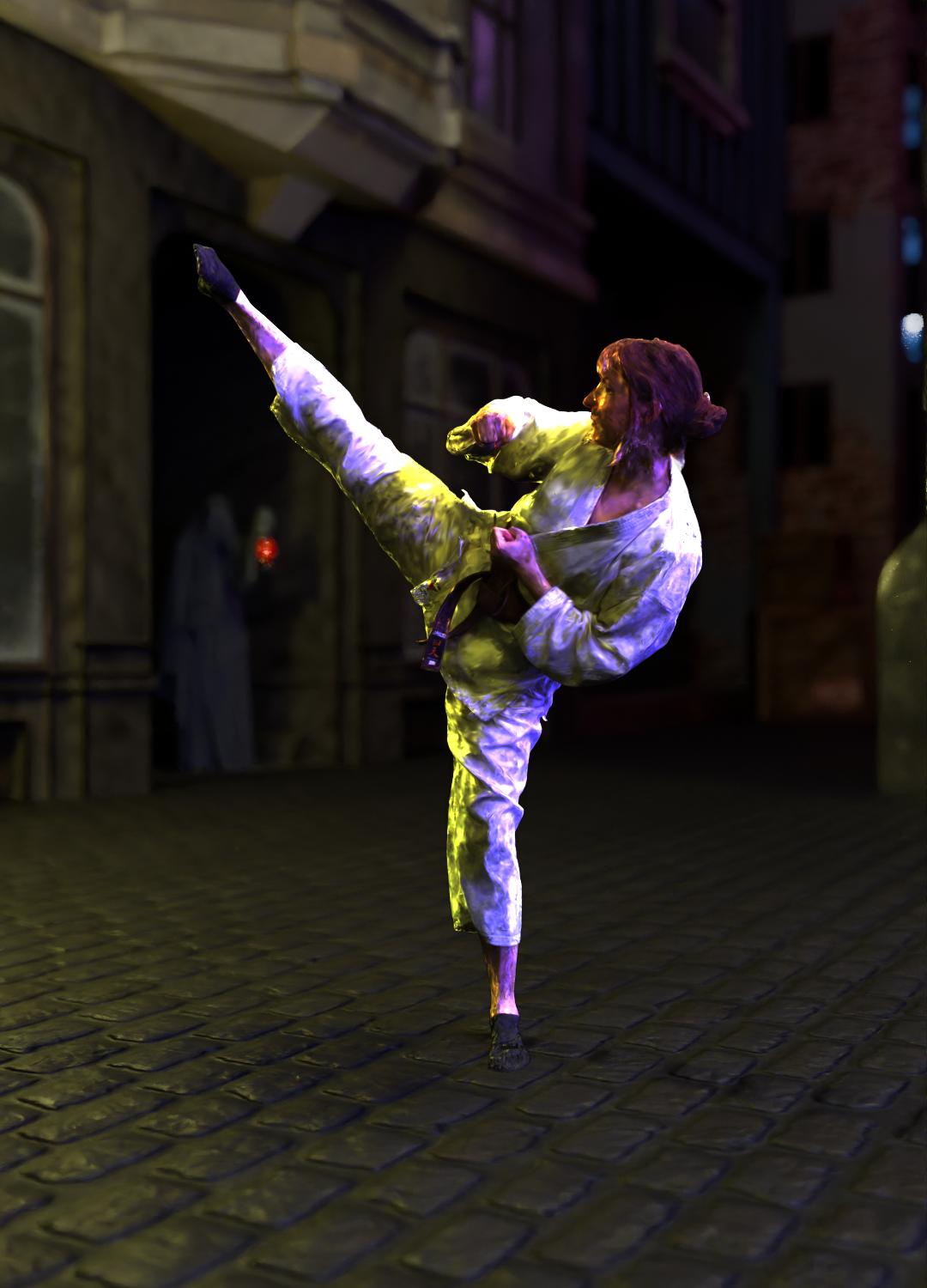}
\end{minipage}
\begin{minipage}[b]{0.16\linewidth}
    \includegraphics[width=\linewidth]{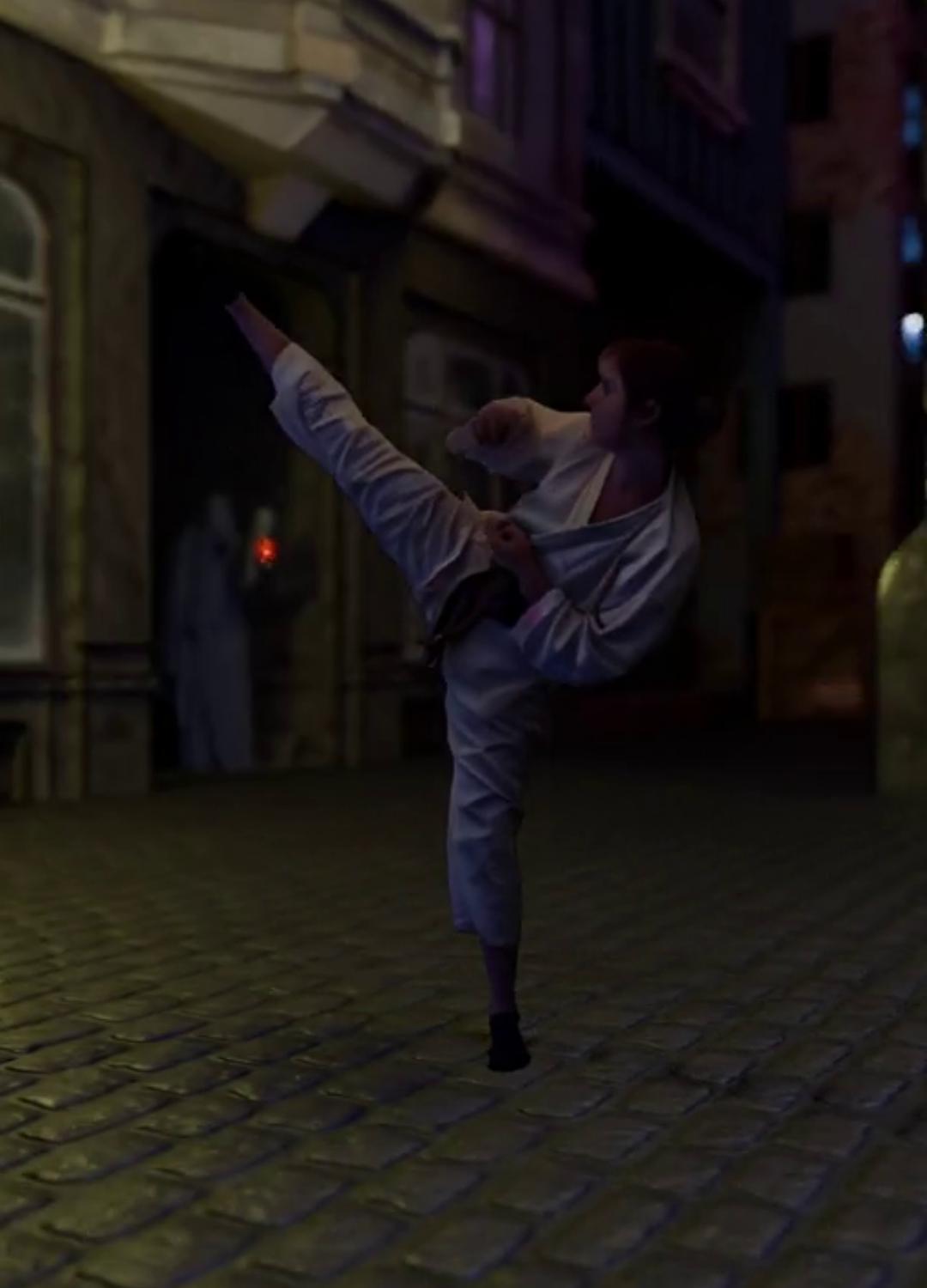}
\end{minipage}
\begin{minipage}[b]{0.16\linewidth}
    \includegraphics[width=\linewidth]{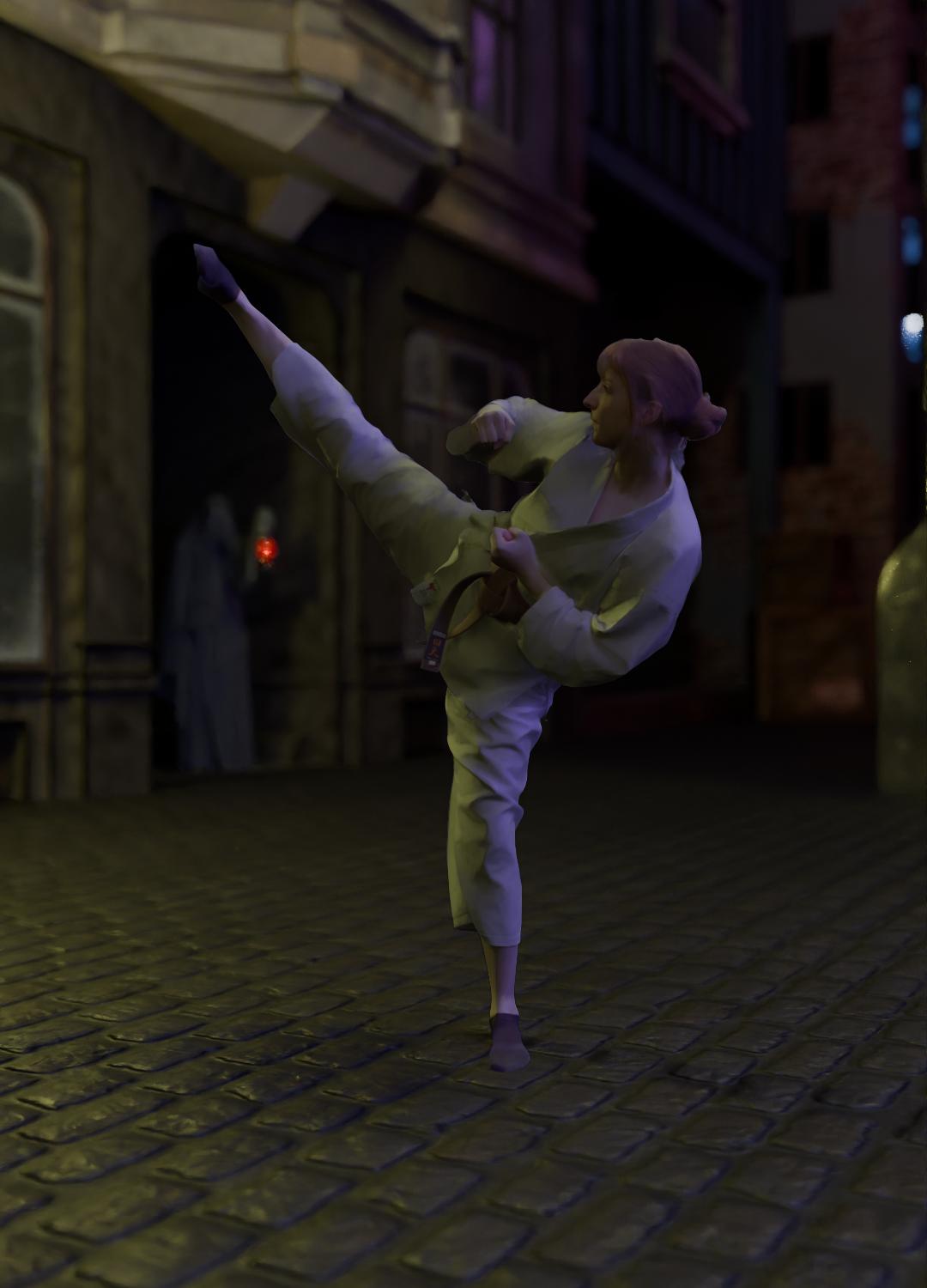}
\end{minipage}

\begin{minipage}[b]{0.16\linewidth}
    \includegraphics[width=\linewidth]{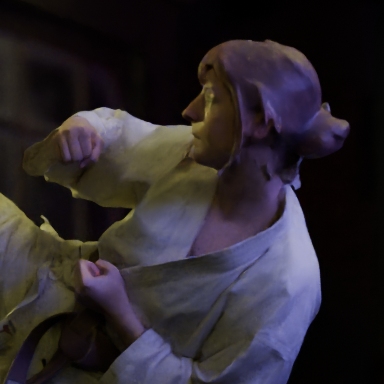}
\end{minipage}
\begin{minipage}[b]{0.16\linewidth}
    \includegraphics[width=\linewidth]{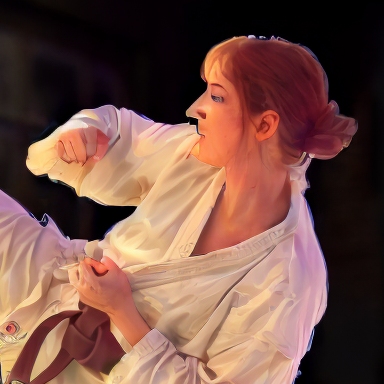}
\end{minipage}
\begin{minipage}[b]{0.16\linewidth}
    \includegraphics[width=\linewidth]{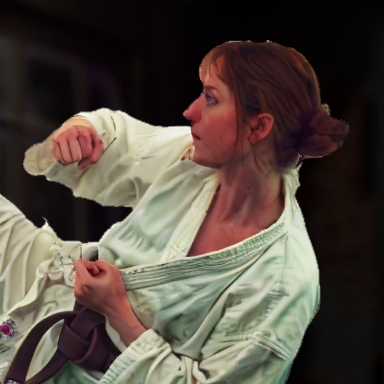}
\end{minipage}
\begin{minipage}[b]{0.16\linewidth}
    \includegraphics[width=\linewidth]{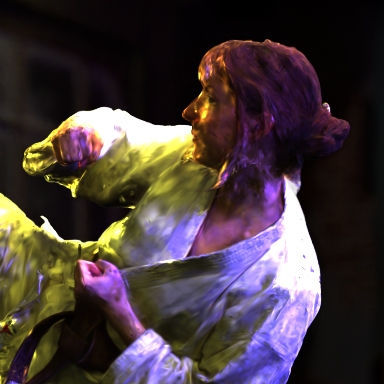}
\end{minipage}
\begin{minipage}[b]{0.16\linewidth}
    \includegraphics[width=\linewidth]{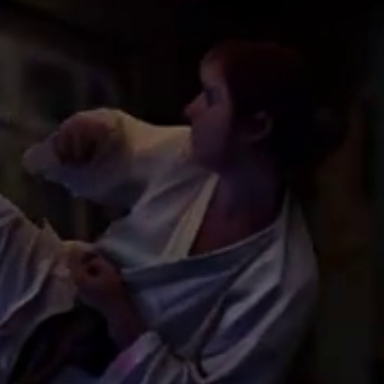}
\end{minipage}
\begin{minipage}[b]{0.16\linewidth}
    \includegraphics[width=\linewidth]{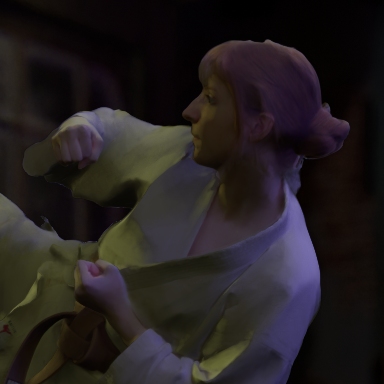}
\end{minipage}

\vspace{1pt}

\begin{minipage}[b]{0.16\linewidth}
    \includegraphics[width=\linewidth]{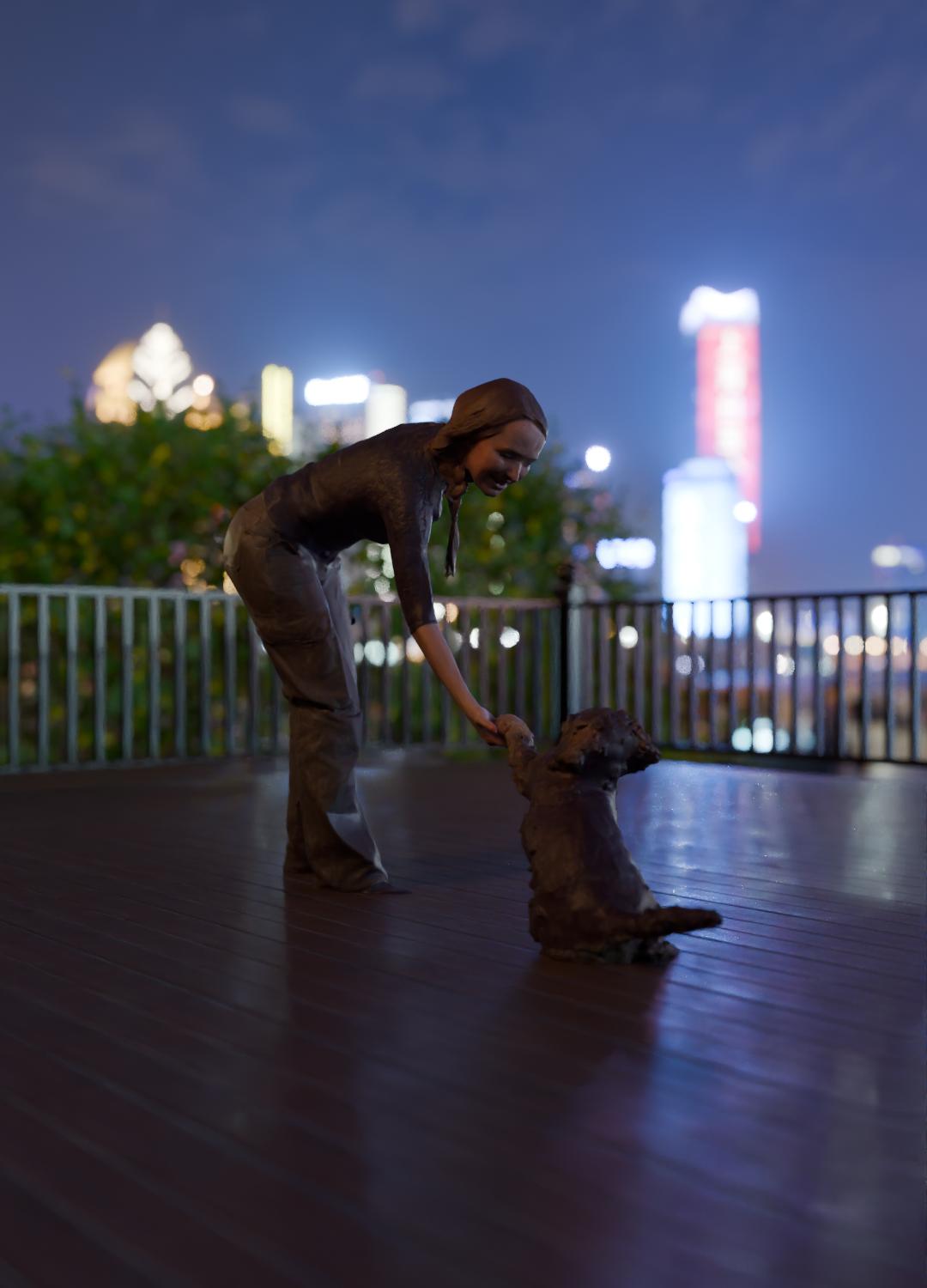}
\end{minipage}
\begin{minipage}[b]{0.16\linewidth}
    \includegraphics[width=\linewidth]{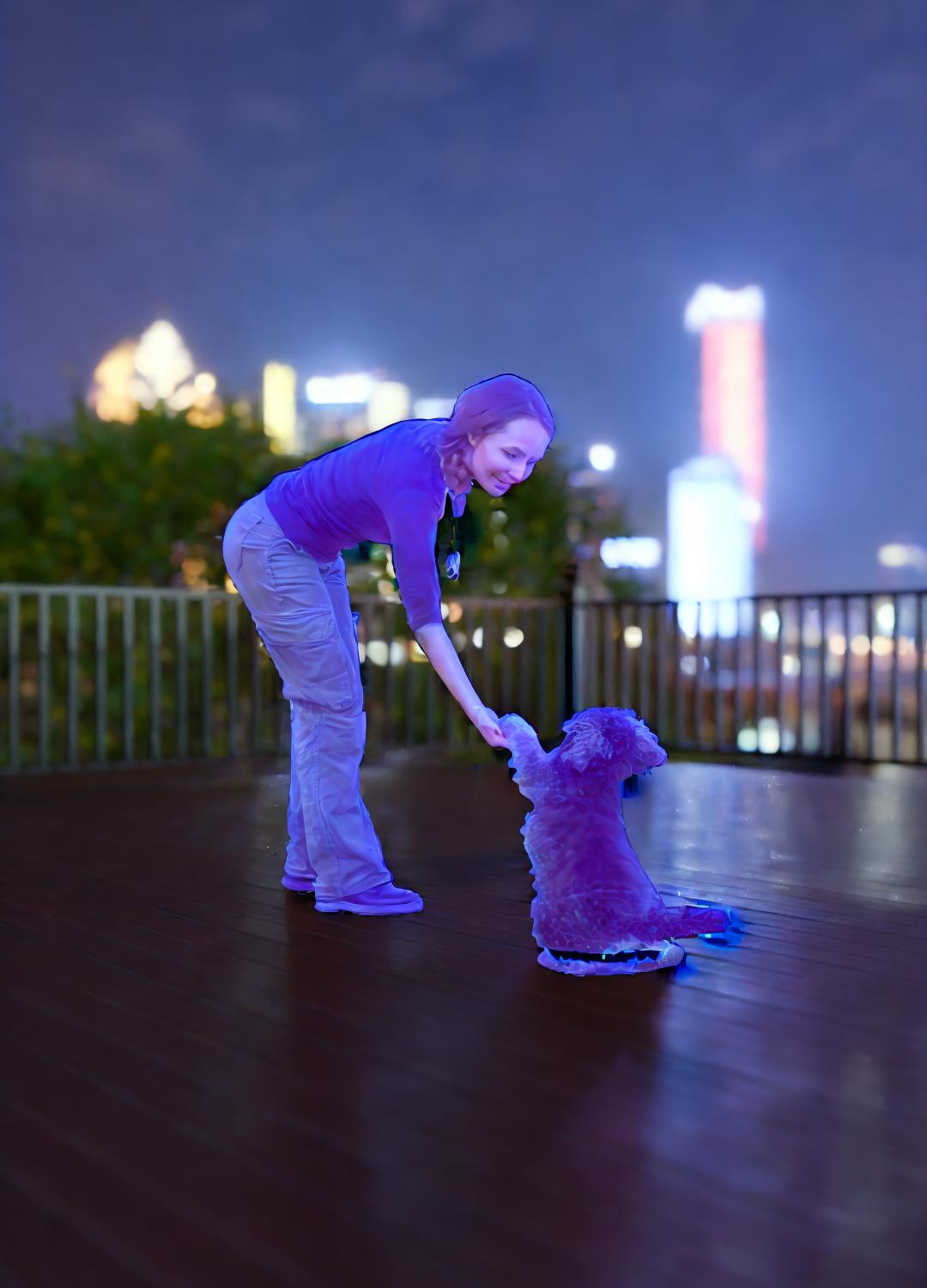}
\end{minipage}
\begin{minipage}[b]{0.16\linewidth}
    \includegraphics[width=\linewidth]{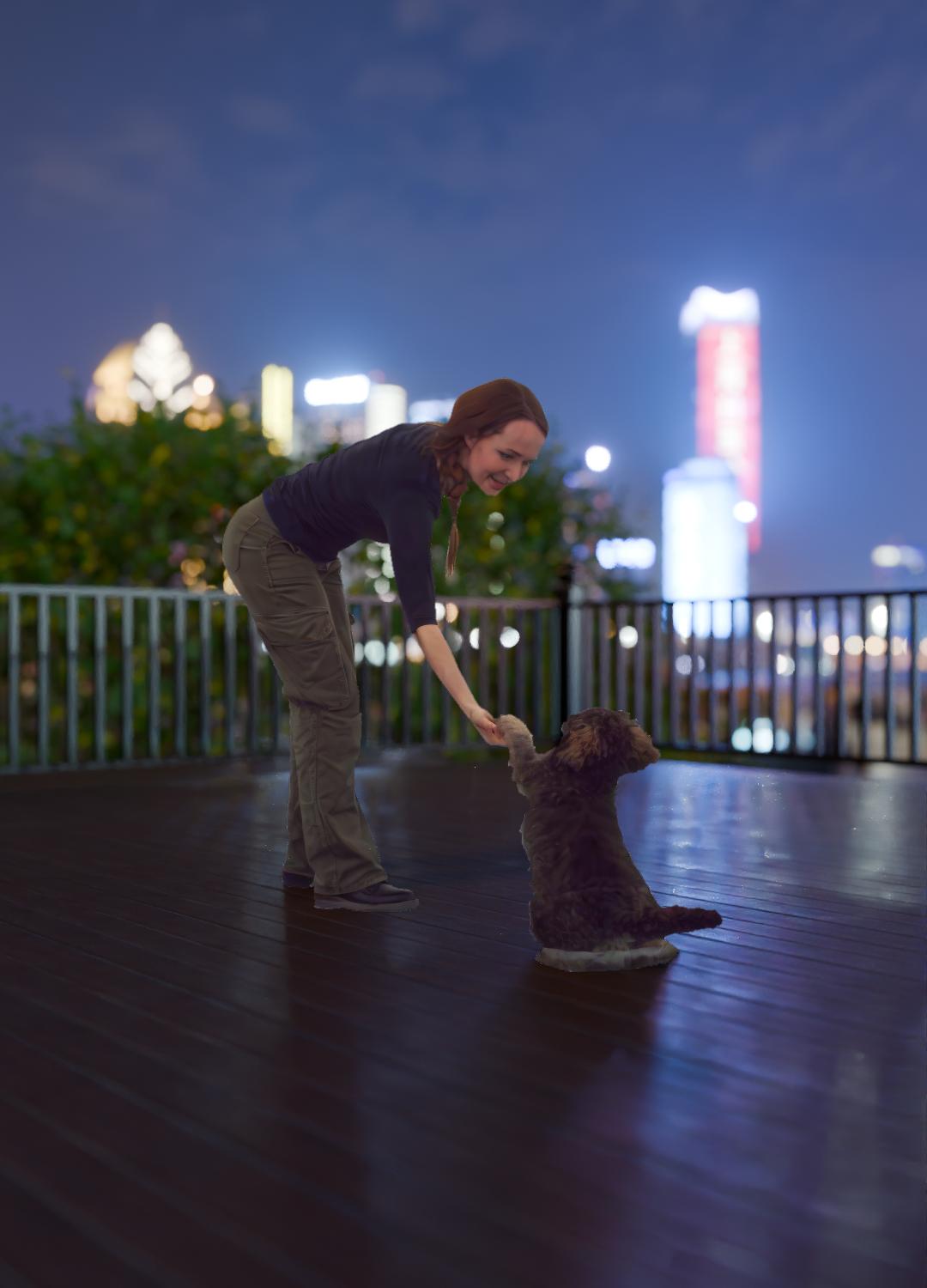}
\end{minipage}
\begin{minipage}[b]{0.16\linewidth}
    \includegraphics[width=\linewidth]{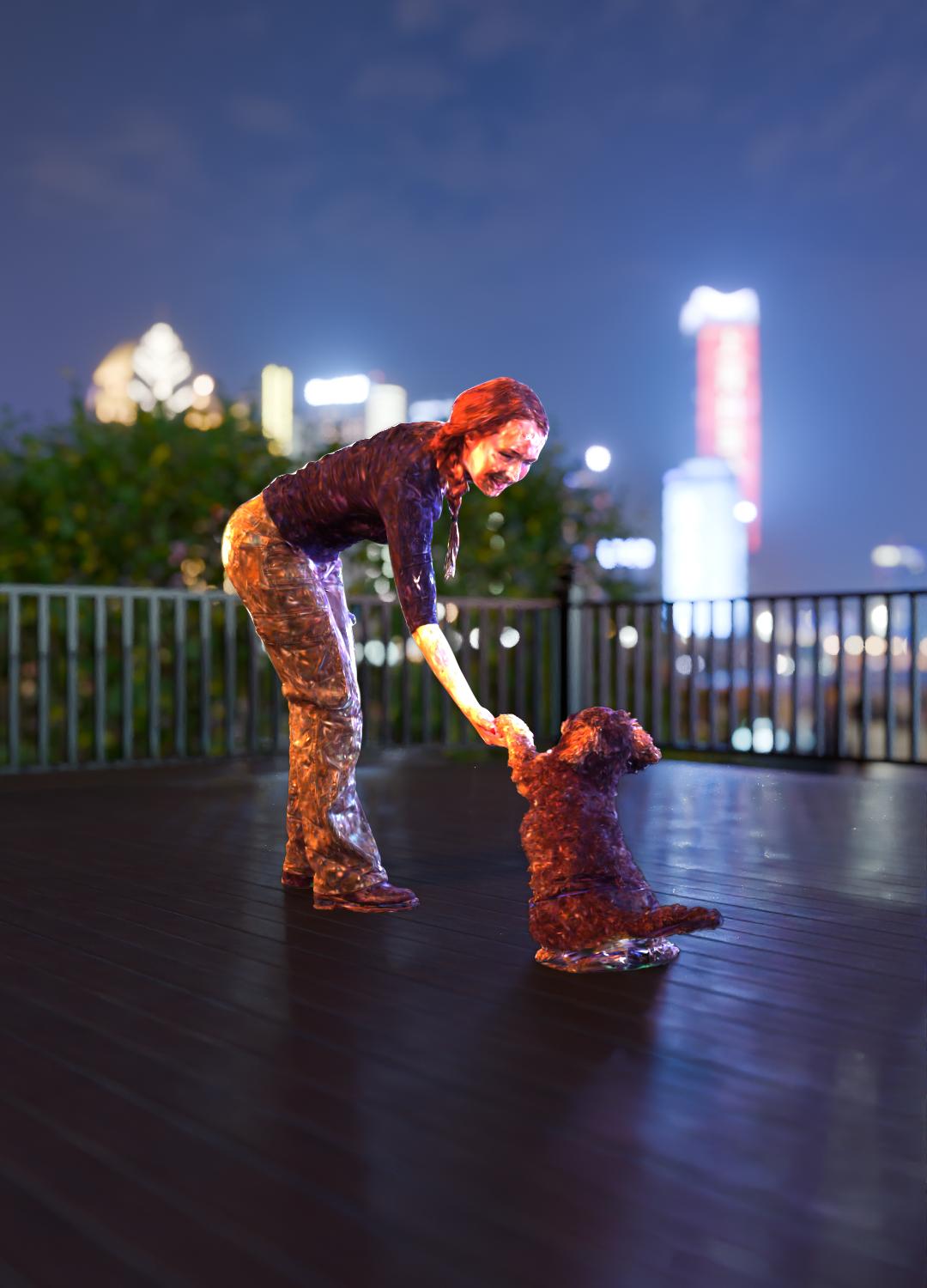}
\end{minipage}
\begin{minipage}[b]{0.16\linewidth}
    \includegraphics[width=\linewidth]{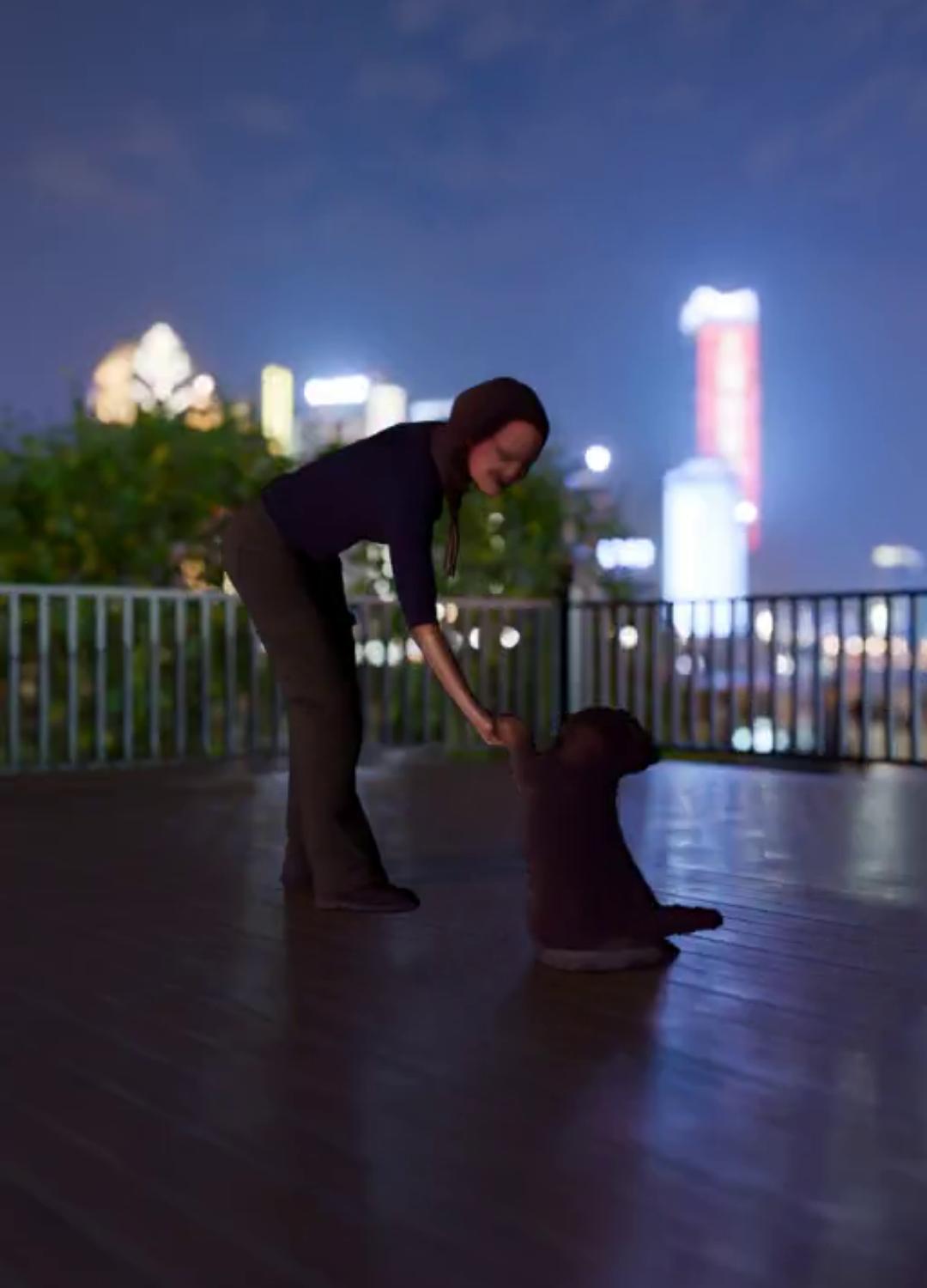}
\end{minipage}
\begin{minipage}[b]{0.16\linewidth}
    \includegraphics[width=\linewidth]{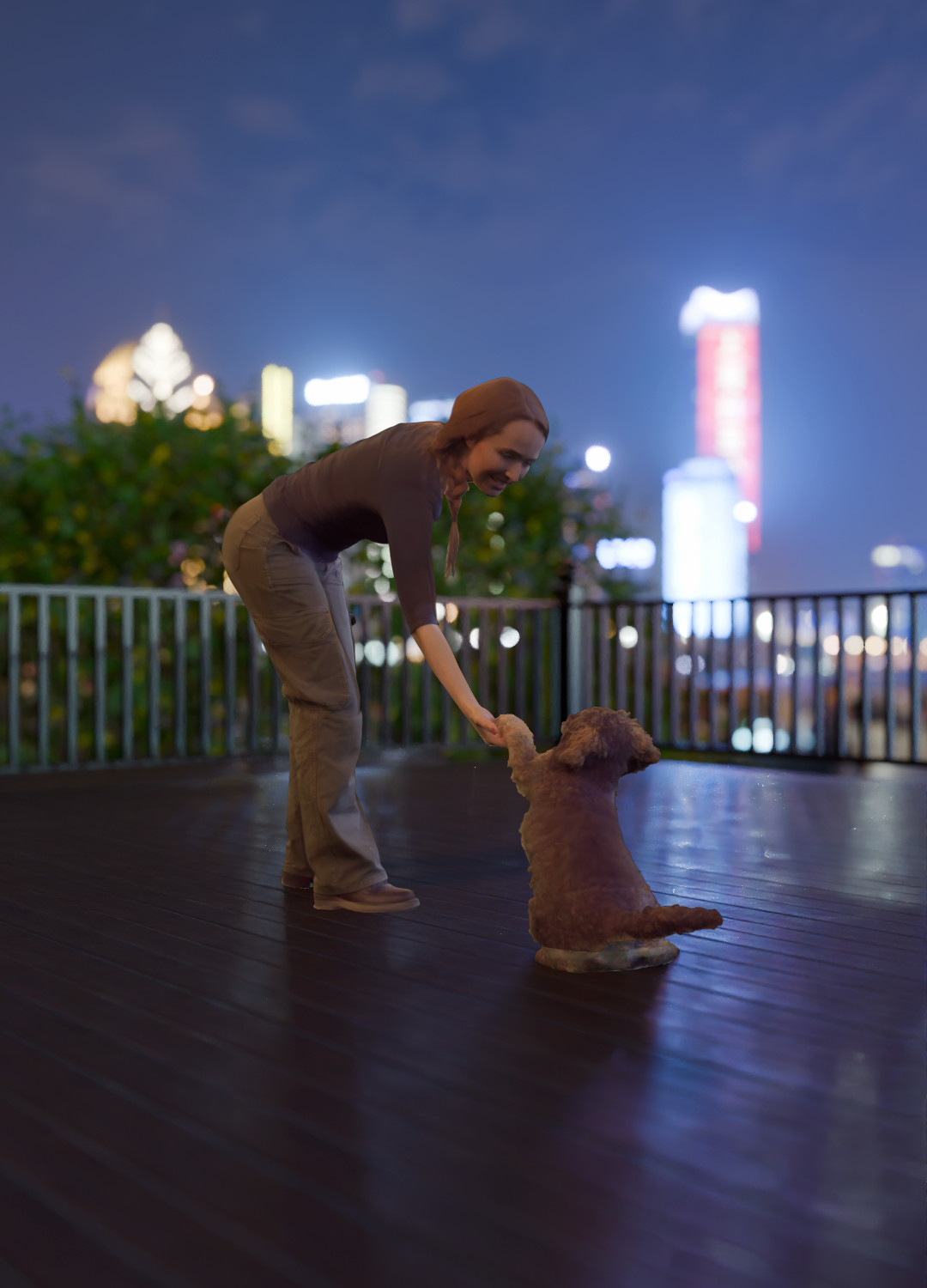}
\end{minipage}

\begin{minipage}[b]{0.16\linewidth}
    \includegraphics[width=\linewidth]{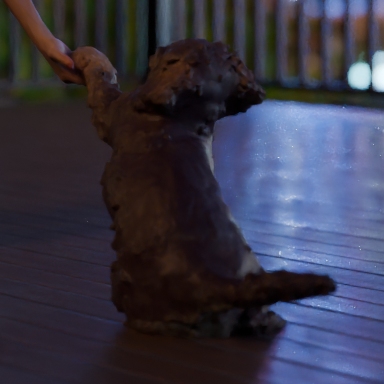}
\end{minipage}
\begin{minipage}[b]{0.16\linewidth}
    \includegraphics[width=\linewidth]{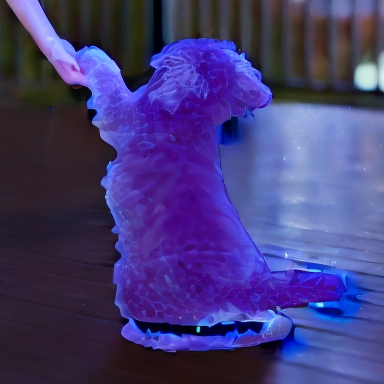}
\end{minipage}
\begin{minipage}[b]{0.16\linewidth}
    \includegraphics[width=\linewidth]{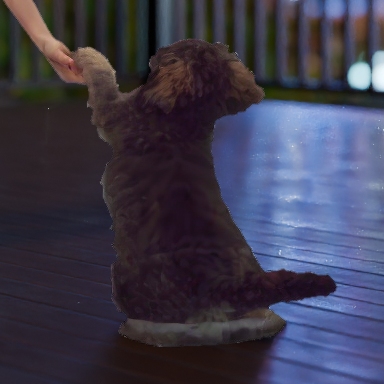}
\end{minipage}
\begin{minipage}[b]{0.16\linewidth}
    \includegraphics[width=\linewidth]{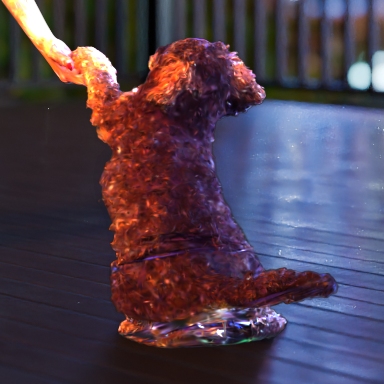}
\end{minipage}
\begin{minipage}[b]{0.16\linewidth}
    \includegraphics[width=\linewidth]{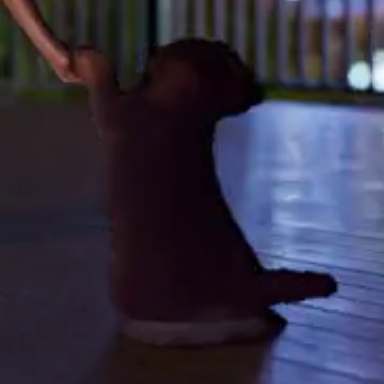}
\end{minipage}
\begin{minipage}[b]{0.16\linewidth}
    \includegraphics[width=\linewidth]{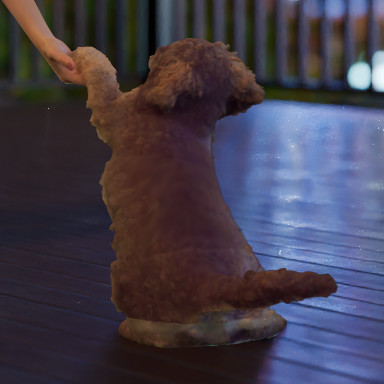}
\end{minipage}

\vspace{2pt}
\scriptsize
\begin{minipage}[b]{0.16\linewidth}
  \centering
  Reference
\end{minipage}
\begin{minipage}[b]{0.16\linewidth}
  \centering
  IC-Light \cite{zhang2025scaling}
\end{minipage}
\begin{minipage}[b]{0.16\linewidth}
  \centering
  Neu.\ Gaffer \cite{jin2024neural}
\end{minipage}
\begin{minipage}[b]{0.16\linewidth}
  \centering
  R3DG \cite{gao2024relightable}
\end{minipage}
\begin{minipage}[b]{0.16\linewidth}
  \centering
  D.\ Render \cite{liang2025diffusion}
\end{minipage}
\begin{minipage}[b]{0.16\linewidth}
  \centering
  Yesnt (Ours)
\end{minipage}

\vspace{-5pt}
\caption{Qualitative comparison with recent state-of-the-art relighting methods. 
Each row corresponds to a different scene under novel lighting conditions. 
Top panels show the full relit view, while bottom panels provide close-up crops.}
\label{fig:qualitative}
\end{figure*}

%% file: figures/alpha/alpha.tex
\begin{wrapfigure}[12]{r}{0.48\linewidth}
\vspace{-25pt}
\vspace{-25pt}
\centering

\begin{minipage}[b]{0.49\linewidth}
    \includegraphics[trim={0 6cm 0 2cm},clip, width=\linewidth]{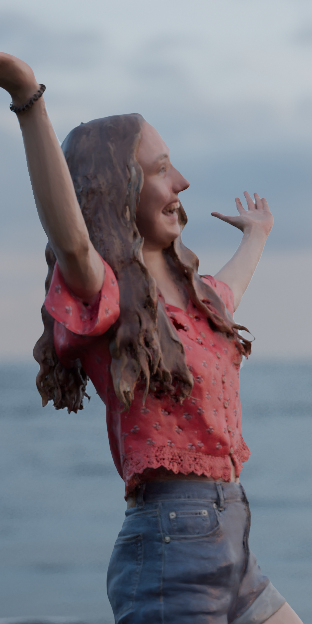}
\end{minipage}
\begin{minipage}[b]{0.49\linewidth}
    \includegraphics[trim={0 6cm 0 2cm},clip, width=\linewidth]{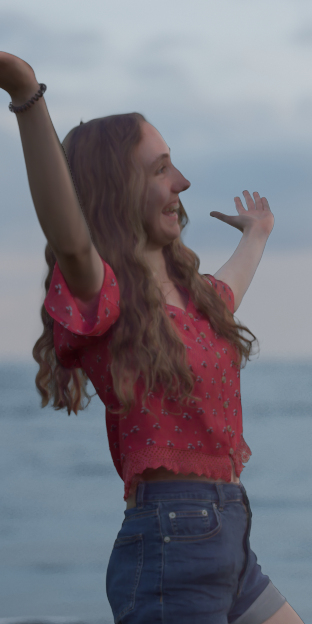}
\end{minipage}

\vspace{-8pt}
\caption{Reference  (left) and our relighting (right), showing smoother surface reconstruction and cleaner alpha mattes with our approach, particularly in challenging regions such as hair.}

\label{fig:alpha}

\end{wrapfigure}

%% file: table/runtimes.tex
\begin{wraptable}[21]{r}{0.48\linewidth}
\vspace{-33pt}
\centering
\caption{Runtimes Analysis of our method compared to others. Bottlenecks are the Gaussian model optimization and the diffusion evaluations.}
\vspace{2pt}
\begin{tabular}{l r}
\toprule
\textbf{Step}       & \textbf{Runtime} \\
\midrule
\multicolumn{2}{c}{\emph{Per Scene}} \\
\midrule
GOF -- Optimization~\cite{yu2024gaussian} & 1.1 h \\
R3DG -- Optimization~\cite{gao2024relightable} & 0.9 h \\
\midrule
\multicolumn{2}{c}{\emph{Per Frame}} \\
\midrule
GOF -- Rendering & 0.4 s \\
Diffusion Decomposition & 80 s \\
Temporal Regularisation & 1.5 s \\
Relighting & 1.5 s \\
Full Pipeline & 83 s \\
\midrule
IC-Light~\cite{zhang2025scaling} & 14 s \\
Neural Gaffer~\cite{jin2024neural} & 18 s \\
R3DG -- Relighting~\cite{gao2024relightable} & 1 s \\
Diffusion Renderer~\cite{liang2025diffusion} & 28 s \\
\bottomrule
\end{tabular}
\label{tab:runtimes}
\end{wraptable}

%% file: sec/5_conclusion.tex
\section{Conclusion}
\label{sec:conclusion}

We asked: \emph{Are diffusion relighting models ready for capture stage compositing?}  
While they produce impressive results, they do not scale well to longer sequences.  
By aggregating stochastic material estimates and enforcing temporal coherence with flow-guided smoothing, our method addresses this shortcoming.  
Experiments on real and synthetic captures show higher fidelity and reduced flickering, bridging the gap between prototypes and production pipelines.  
Our results suggest that enforcing structured temporal constraints is more effective than increasing generative model capacity for achieving stable video relighting.
In short, the answer is \emph{Yesnt}: diffusion models alone are not yet ready, but our hybrid approach offers a practical path toward production-ready volumetric video relighting.

\section*{Acknowledgments}
This work has been funded by the Ministry of Culture and Science North Rhine-Westphalia under grant number PB22-063A (InVirtuo 4.0: Experimental Research in Virtual Environments).